\documentclass[nonatbib]{article}

\usepackage[preprint]{stylefile} 
\usepackage[utf8]{inputenc} 
\usepackage[T1]{fontenc}    
\usepackage{hyperref}       
\usepackage{url}            
\usepackage{booktabs}       
\usepackage{amsfonts}       
\usepackage{nicefrac}       
\usepackage{microtype}      
\usepackage{float}         

\usepackage{pdflscape}
\usepackage{longtable}
\usepackage{booktabs}
\newif\ifshowskeleton
\showskeletontrue 
\usepackage{wrapfig}   
\usepackage{xcolor}
\usepackage{makecell,colortbl,siunitx}
\definecolor{myred}   {HTML}{F2A29B}  
\definecolor{myorange}{HTML}{F6C991}
\definecolor{myyellow}{HTML}{F6E79B}
\definecolor{mylgreen}{HTML}{C7E6AC}
\definecolor{mydgreen}{HTML}{9ACF78}  
\usepackage{adjustbox}

\definecolor{cb0}{HTML}{ff9999}  
\definecolor{cb1}{HTML}{ffcc99}  
\definecolor{cb2}{HTML}{ffe699}  
\definecolor{cb3}{HTML}{ccffcc}  
\definecolor{cb4}{HTML}{99e699}  
\definecolor{cb5}{HTML}{66cc66}  

\usepackage{siunitx}  

\newcommand{\heatcellC}[1]{%
  \begingroup
  \def\val{#1}%
  \ifdim\val pt<0pt
    \cellcolor{cb0}{$\num[round-precision=2]{\val}$}%
  \else
    \ifdim\val pt<0.20pt \cellcolor{cb1}{+\num[round-precision=2]{\val}}%
    \else\ifdim\val pt<0.35pt \cellcolor{cb2}{+\num[round-precision=2]{\val}}%
    \else\ifdim\val pt<0.50pt \cellcolor{cb3}{+\num[round-precision=2]{\val}}%
    \else\ifdim\val pt<0.75pt \cellcolor{cb4}{+\num[round-precision=2]{\val}}%
    \else                 \cellcolor{cb5}{+\num[round-precision=2]{\val}}%
    \fi\fi\fi\fi
  \fi
  \endgroup
}

\newcommand{\scorecell}[2]{%
  \cellcolor{#1}\makebox[1.4cm][c]{\vphantom{0.0}#2}%
}

\newcommand{\bucket}[1]{
  \ifdim #1 pt <0.30pt \color{myred}   #1\else
  \ifdim #1 pt <0.50pt \color{myorange}#1\else
  \ifdim #1 pt <0.70pt \color{myyellow}#1\else
  \ifdim #1 pt <0.85pt \color{mylgreen}#1\else
                     \color{mydgreen}#1\fi\fi\fi\fi}

\newcommand{\heatcellB}[1]{%
  \begingroup
    \edef\val{#1}%
    \ifdim\val pt<0.30pt  \cellcolor{myred}{\num{\val}}%
    \else\ifdim\val pt<0.50pt \cellcolor{myorange}{\num{\val}}%
    \else\ifdim\val pt<0.70pt \cellcolor{myyellow}{\num{\val}}%
    \else\ifdim\val pt<0.85pt \cellcolor{mylgreen}{\num{\val}}%
    \else                 \cellcolor{mydgreen}{\num{\val}}%
    \fi\fi\fi\fi
  \endgroup}

\usepackage{etoolbox}        
\makeatletter
  \let\origmidrule\midrule      
\makeatother    

\makeatletter
\pretocmd{\midrule}{\arrayrulecolor{gray!60}}{}{}
\apptocmd{\midrule}{\arrayrulecolor{black}}{}{}
\makeatother

\renewcommand{\paragraph}[1]{%
  \noindent\textbf{#1}%
}

\newcommand{\eg}{\textit{eg. }}
\newcommand{\ie}{\textit{ie. }}

\newcommand{\quotes}[1]{``#1''}

\usepackage{makecell}
\usepackage[most]{tcolorbox}[colback=blue!10, colframe=black]
\usepackage{multirow}
\usepackage{graphicx}
\usepackage{subcaption}
\usepackage{amsmath}
\usepackage{wrapfig}
\usepackage{amsthm}
\usepackage{cleveref}
\crefname{figure}{Figure}{Figures}
\crefname{section}{Section}{Sections}
\crefname{table}{Table}{Tables}
\newtheorem{definition}{Definition}

\usepackage[style=numeric]{biblatex}

\AtEveryBibitem{%
  \clearfield{urldate}%
  \clearlist{urldate}%
}
\DeclareFieldFormat{urldate}{}       
\renewbibmacro*{urldate}{}           

\title{On the rankability of visual embeddings}

\author{
Ankit Sonthalia$^{1}$\thanks{Corresponding author: \texttt{ankit.sonthalia@uni-tuebingen.de}} \quad
Arnas Uselis$^{1}$ \quad
Seong Joon Oh$^{1}$ \\
$^{1}$Tübingen AI Center, Universität Tübingen, Germany \\
}

\begin{document}

\maketitle

\begin{abstract}
We study whether visual embedding models capture continuous, ordinal attributes along linear directions, which we term \textit{rank axes}. We define a model as \textit{rankable} for an attribute if projecting embeddings onto such an axis preserves the attribute's order. Across 7 popular encoders and 9 datasets with attributes like age, crowd count, head pose, aesthetics, and recency, we find that many embeddings are inherently rankable. Surprisingly, a small number of samples, or even just two extreme examples, often suffice to recover meaningful rank axes, without full-scale supervision. These findings open up new use cases for image ranking in vector databases and motivate further study into the structure and learning of rankable embeddings. Our code is available at \href{https://github.com/aktsonthalia/rankable-vision-embeddings}{https://github.com/aktsonthalia/rankable-vision-embeddings}.
\end{abstract}

\section{Introduction}

Visual embeddings are widely used for image retrieval. This relies on embeddings forming a metric space, where similar images are placed nearby. Modern visual encoders generally satisfy this property, and many systems depend on it in the form of vector databases.

Ranking is another core operation in databases. It allows users to navigate large collections by sorting, paginating, and filtering results. For instance, e-commerce platforms like Amazon benefit from ranking product images by visual quality or certain product-specific attributes (\eg ranking shoes by how formal they look).

\begin{wrapfigure}{r}{.5\linewidth}
    \vspace{-1em}
    \includegraphics[width=\linewidth]{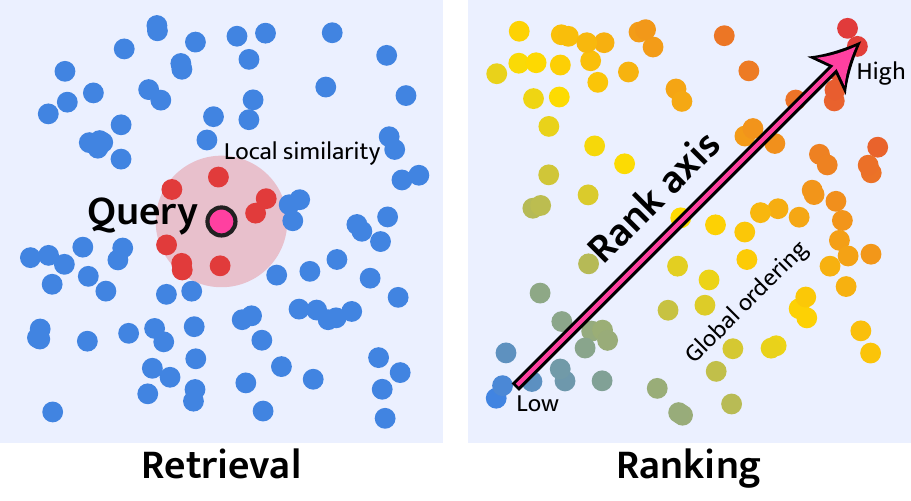}
    \vspace{-3em}
\end{wrapfigure}

In this work, we ask: \textbf{are visual embeddings also rankable?} Retrieval relies on \textit{local} similarity around a query. Ranking requires a \textit{global} ordering along an attribute. Prior work has largely addressed the former. The global rankability of embeddings remains underexplored.

We define \textbf{rankability} as follows: given an embedding function $f$ and a continuous attribute $A$ (e.g. ``age''), we say $f$ is rankable with respect to $A$ if there exists a \textbf{rank axis} $v_A$ such that the projection  $v_A^\top f(x)$ preserves the correct order of the target attribute $A(x)$ over a dataset. For instance, if $A$ denotes \quotes{age}, this projection would sort face images from youngest to oldest.

We examine two questions: (1) Are visual embeddings rankable? (2) How easily can we recover the rank axis for a given attribute?

To address (1), we evaluate 7 modern visual encoders, from ResNet to CLIP, across 9 datasets with 7 attributes: age, crowd count, 3 head pose angles (pitch, roll, yaw), image aesthetics, and recency. We find that many embedding spaces are indeed rankable (\cref{sec:are_embeddings_rankable}). 

To address (2), we estimate the rank axis $v_A$ with minimal supervision. The structure of the embedding space makes full-dataset regression unnecessary. In many cases, a handful of annotated samples and, in some cases, a pair of samples $x_l$ (low) and $x_h$ (high) already recover non-trivial ranking performance. For the latter case, we define the rank axis as $v_A = (f(x_h) - f(x_l))/\|f(x_h) - f(x_l)\|_2$. This opens up the possibility for fast ordering of new images by arbitrary attributes. For example, a photo app lets users sort selfies by age appearance. It uses CLIP embeddings and two reference images: one of a child and one of an elderly person. The app computes $v_{\text{age}}$ without training. Users scroll from youngest-looking to oldest-looking faces in their album (\cref{sec:finding-rank-axis}).

Contributions:
\begin{enumerate}
    \item We define and motivate rankability as a property of visual embeddings, distinct from retrieval.
    \item We study rankability across modern encoders and real-valued attributes; results show that current embeddings are rankable.
    \item We show that rank axes can sometimes be recovered using only two or a handful of labelled samples.
\end{enumerate}

\section{Related work}

\paragraph{Embeddings for retrieval.}  
Visual encoders are commonly used to index images in vector databases, enabling nearest neighbour search for retrieval tasks~\cite{musgrave2020metric,Radford2021LearningTVA,li2022blip,schall2023imageretrieval}.
This setup, known as deep metric learning~\cite{Chen2021DeepLFA,chen2022deep,musgrave2020metric}, predates vision-language models like CLIP~\cite{Radford2021LearningTVA}.
CLIP and related models shifted the focus to \textit{cross-modal} similarity modelling, where vision and language share a joint embedding space used for classification~\cite{Radford2021LearningTVA}, retrieval~\cite{yao2021filip,Baldrati2022ConditionedACA}, and captioning~\cite{mokady2021clipcap,Kuo2022BeyondAPA}.
While the majority of work in visual encoders is devoted to the understanding of the local similarity structure, we study how visual embeddings support \textit{global} ranking instead of just local retrieval.

\paragraph{Improving embedding geometry and structure.}  
Prior work has explored ways to improve the geometry of the embedding space.
Order embeddings and hyperbolic representations have been used to model hierarchies~\cite{vendrov2015orderembeddings, khrulkov2019hyperbolic,Desai2023HyperbolicIRA,Qiao2024HYDENHDA}.
Training disentangled representation~\cite{wang2024disentangled} is considered critical for compositionality, where attributes are assigned to certain linear subspaces~\cite{trager2023linearsoa, berasi2025notota}.
Others have defined concepts like uniformity and separability of the representations~\cite{wang2020understandingcra}.
In this work, we focus on the analysis of a wide range of visual encoders, rather than introducing recipes for improvements.

\paragraph{Analysing embedding geometry and structure.}  
A large body of work has examined the geometry and structure of CLIP's learned embedding space.  
CLIP and its derivatives have been studied extensively~\cite{chen2020simple,radford2021learning,li2021align,yu2022coca}.
Several works have reported modality gaps between vision and language embeddings~\cite{fahim2024itsnaa}.  
Some studies point to the absence of certain structures and capabilities in CLIP representations: attribute-object bindings~\cite{lewis2022does,yuksekgonul2022and,darina2025binding}, or the association of attributes to corresponding instances.
Others argue that much information is already present in CLIP representations, including parts-of-speech and linguistic structure~\cite{oldfield2023partsofsa}, attribute-object bindings~\cite{koishigarina2025clip}, and compositional attributes~\cite{uselis2025decodability}. The platonic representation hypothesis further suggests that models converge to similar internal structures~\cite{huh2024platonic}. 
In this work, we analyse the embedding geometry and structure for modern visual embeddings from the novel perspective of rankability.

\paragraph{Linearly probing an embedding.}  
Linear probing is a fast and widely used method to test for the presence of concepts in visual embeddings~\cite{kim2018interpretability,huang2024lppp,arnas2025iclr}.  
It measures the accuracy of a linear classifier trained on intermediate-layer features, effectively testing whether a hyperplane can separate embeddings containing a concept from those that do not.  
This technique has been used to study the geometry of CLIP's embeddings~\cite{levi2024doubleellipsoid} and to probe for specific information such as attribute-object bindings and compositionality~\cite{koishigarina2025clip,uselis2025decodability}.  
While effective for binary separability, linear probes are limited in capturing non-binary, ordinal, or relational structures~\cite{alain2016understanding}.  
Prior work has not directly analysed how \textit{continuous} attributes are laid out in the embedding space. 
Our work extends this line of research by moving beyond concept detection to characterise the internal structure of embeddings along ordinal axes, introducing rankability as a new property not captured by prior probing methods.

\paragraph{Ordinal information in embeddings.}  
Early works on relative and ordinal attributes explored how continuous visual attributes could be inferred and ranked~\cite{parikh2011relative,liang2014beyond,Zoran_2015_ICCV,han2017heterogeneous,yu2015just}.  
However, these efforts were limited to smaller models and datasets.  
Recent studies have begun to examine ordinal signals in large pre-trained models.  
These include aligning CLIP with ordinal supervision~\cite{wang2023learning}, and applying CLIP or general VLMs to specific tasks such as aesthetics~\cite{xu2023clip,wang2023exploring}, object counting \cite{paiss2023teaching, jiang2023clipcount}, crowd counting~\cite{liang2023crowdclip}, and difference detection~\cite{sam2024finetuning, kil2024compbench}. 
Several works have adapted CLIP for ranking through prompt tuning~\cite{li2022ordinalclip}, adapter-based methods~\cite{yu2024rankingaware} or regression-based fine-tuning~\cite{du2024teach,wang2023learningtorank}.  
Despite these efforts, most focus on task-specific performance or modifying the embedding space, rather than understanding its internal ordinal structure in the embeddings themselves.  
Our work fills this gap by systematically analysing the rankable structure of existing visual encoders and revealing the presence of ordinal directions in their embedding spaces.

\section{Are vision embeddings rankable?}
\label{sec:are_embeddings_rankable}

We set out to answer the question. For this, we first formally define rankability and strategies to measure it (\cref{sec:defining_rankability}). 
We introduce the models and data used for our experiments in \cref{sec:experimental_details}. We present results in \cref{sec:general_results}.

\paragraph{Notation.}
Our experiments use RGB image datasets with real-valued attribute labels. We denote an image dataset as $X \subset \mathbb{R}^{3 \times H \times W}$ and define an \textit{ordinal attribute} $A$ as a function $A: X \rightarrow Y \subset \mathbb{R}$ where $Y$ is the range of possible labels and $A(x)$ is the ground-truth label for a given image $x$. 
An image encoder is a function $f: X \rightarrow \mathbb{R}^d$ where $d$ is the dimensionality of the embedding space. We occasionally use the term \quotes{representation} to refer to an image encoder.

\subsection{Rankability}
\label{sec:defining_rankability}

We aim to characterize the \textit{linear} structure of the ordinal information present in visual embedding spaces. Our definition of rankability then naturally emerges as follows.

\begin{definition}[Rankability]
A representation $f: X \rightarrow \mathbb{R}^d$ is \textbf{rankable} for an ordinal attribute $A$ over an image dataset $X$ if there exists a \textbf{rank axis} $v_A \in \mathcal{S}^{d-1}$, a $d-1$ dimensional unit sphere, such that for any $x_1, x_2 \in X$ with $A(x_1) \geq A(x_2)$, it follows that $v_A^\top f(x_1) \geq v_A^\top f(x_2)$. 
\end{definition}

The above definition requires a rank axis $v_A$ to exactly preserve the ordering provided by the attribute $A$ over the dataset $X$. In practice, we measure rankability using the generalisation performance of the rank axis $v_A$ learned on a training split $X_\text{train}$ and tested on a disjoint split $X_\text{test}$. In this section, we obtain $v_A$ via linear regression on the labelled samples: $\{(f(x_i),a_i)\}_i$, where $a_i\in\mathbb{R}$ is the ground-truth continuous attribute label for each $x_i$. In \cref{sec:finding-rank-axis}, we consider approaches that do not require access to the attribute labels $a_i$.

\textbf{Spearman's rank correlation coefficient (SRCC)}, denoted as $\rho$ serves as our primary metric quantifying the monotonicity of the relationship between the true ordinality of the attribute $A$ and the predicted ranking along $v_A$. 
SRCC is widely used in related contexts \cite{yu2025rankingaware}.

We provide three reference points for the obtained rankability to contextualise the SRCC values:

\textbf{(1) No-train (``lower bound'').} Even for untrained visual encoders, the embeddings do not result in null rank correlation ($\rho$=0). In order to correctly capture the no-information baseline, we consider the performances of randomly initialized version of each encoder considered \cite{ulyanov2018deep}. The optimal rank axis $v_A$ in this space serves as a lower bound for the rankability of the pretrained encoder.  

\textbf{(2) Nonlinear (upper bound for embedding).} To estimate the total ordinal information in the given embedding, we use a two-layer multilayer perceptron (MLP), known to be a universal approximator \cite{hornik1989multilayer}. Comparing against a non-linear regressor lets us estimate the proportion of ordinal information in an embedding that can be extracted linearly with a rank axis.  

\textbf{(3) Finetuning (upper bound for encoder architecture).} To measure a broader upper bound indicating the capacity of the encoder architecture and the learnability of the attribute, we finetune the encoder. This conceptually envelops the nonlinear regression upper bound.

\subsection{Experimental details}
\label{sec:experimental_details}

We provide further details on the list of attributes, datasets, encoder architectures, and the model selection protocol.

\subsubsection{Attributes and datasets}

In total, we use 9 datasets, covering 7 attributes. We provide a detailed breakdown in \cref{tab:dataset-summary}. 

\vspace{-1em}
\begin{table}[H]
\caption{\textbf{Datasets and attributes}. Summary of datasets used for evaluating the rankability of visual representations.}
\vspace{0.5em}
\centering
\resizebox{\linewidth}{!}{%
\begin{tabular}{llllll}
\toprule
\textbf{Attribute} & \textbf{Dataset} & \textbf{\#Train-val} & \textbf{\#Test} & \textbf{Label Range} & \textbf{Split} \\
\origmidrule
\multirow{2}{*}{Age} 
    & UTKFace \cite{zhang2017age} & 13,146 & 3,287 & 21--60 & From \cite{kuprashevich2023mivolo} \\
    & Adience \cite{eidinger2014age} & $\sim$14k & $\sim$4k & 8 age groups & Official 5-fold \\
\midrule
\multirow{3}{*}{Crowd count} 
    & UCF-QNRF \cite{idrees2018composition} & 1,201 & 334 & 49--12,865 & Official \\
    & ShanghaiTech-A \cite{zhang2016singleimage} & 300 & 182 & 33--3,139 & Official \\
    & ShanghaiTech-B \cite{zhang2016singleimage} & 400 & 316 & 9--578 & Official \\
\midrule
Pitch & BIWI Kinect \cite{fanelli2011real} & 10,493 & 4,531 & $\pm 60^\circ$ & 6 test seqs. \\
Yaw   & BIWI Kinect & 10,493 & 4,531 & $\pm 75^\circ$ & 6 test seqs. \\
Roll  & BIWI Kinect & 10,493 & 4,531 & $\pm 45^\circ$ & 6 test seqs. \\
\midrule
\multirow{2}{*}{Aesthetics} 
    & AVA \cite{murray2012ava} & 230,686 & 4,692 & Ratings (1--10) & From \cite{yu2025rankingaware} \\ 
    & KonIQ-10k \cite{hosu2020koniq10k} & 8,058 & 2,015 & Ratings (1--100) & Official \\
\midrule
Image modernness 
    & Historical Color Images \cite{palermo2012dating} & 1,060 & 265 & 5 decades & From \cite{yu2025rankingaware} \\
\bottomrule
\end{tabular}%
}
\vspace{-0.5em}
\label{tab:dataset-summary}
\end{table}

\subsubsection{Architectures}
\label{sec:architectures}

We test representative image-only and CLIP-based encoders. Among image-only encoders, we use the ResNet50 \cite{he2016deep}, ViT-B/32@224px \cite{dosovitskiy_image_2021}, and ConvNeXtv2-L \cite{woo2023convnext} architectures. Likewise, among CLIP encoders, we test the ResNet50, ViT-B/32, and ConvNeXt-L@320px variants. We also test DINOv2 \cite{oquab_dinov2_2024} (ViT-B/14 variant). See \cref{tab:model_overview} for more information.

\subsubsection{Model selection}
\label{sec:model_selection}

Hyperparameters for reporting final performances on the test set are selected based on the best validation SRCC, using either official validation splits or holding them out from the corresponding training splits.

\paragraph{Linear and nonlinear regression.}
We test $30$ random hyperparameter configurations per dataset-model pair. The initial learning rate is sampled from a log-uniform distribution over $[10^{-6},\; 10^{-1}]$ and decayed over a cosine schedule to zero, while the weight decay is sampled from a log-uniform distribution over $[10^{-7}, 10^{-4}]$. Data augmentation (horizontal flipping) is also toggled on or off randomly for a given run. We use $100$ epochs and a batch size of $128$ throughout.

\paragraph{Finetuning.} For image‑only ResNet50 and ConvNeXt encoders, we conduct a grid search over two encoder learning rates ($10^{-4}, 10^{-3}$) and two weight‑decay rates ($0, 10^{-5}$). For ViT‑B/32, DINOv2 and CLIP models, we conduct a larger grid search over three encoder learning rates ($10^{-7}, 10^{-6}, 10^{-5}$) and four weight‑decay values ($0, 10^{-5}, 10^{-4}, 10^{-3}$). The downstream model always uses a learning rate of $0.1$. We use $20$ epochs for all runs. Batch size is fixed at $128$, except for ConvNeXt‑v2, DINOv2 and CLIP‑ConvNeXt‑v2, where batch size is reduced to $16$ because of memory constraints. We use horizontal-flip augmentation in all finetuning runs to mitigate overfitting.

\subsubsection{Compute resources}
\label{sec:compute}

All experiments were conducted on a single NVIDIA A100 GPU with 40GB of memory. All primary experiments using frozen embeddings completed within 1–2 minutes, as we cache the embeddings. Finetuning experiments took between 1 and 24 hours each, depending on the dataset. In total, our experiments amounted to approximately 150 compute-days.

\subsection{Results}
\label{sec:general_results}

\begin{figure}[H]
  \centering

  \begin{subfigure}[t]{0.49\linewidth}
    \centering
    \includegraphics[width=\linewidth]{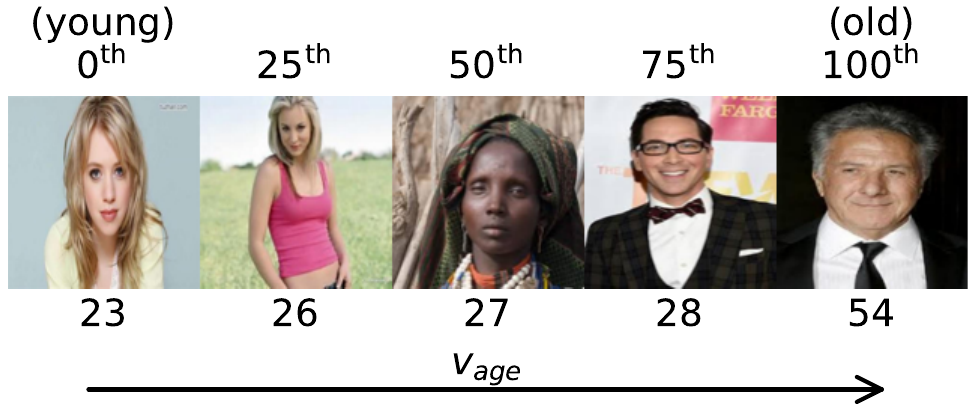}
  \end{subfigure}
  \hfill
  \begin{subfigure}[t]{0.49\linewidth}
    \centering
    \includegraphics[width=\linewidth]{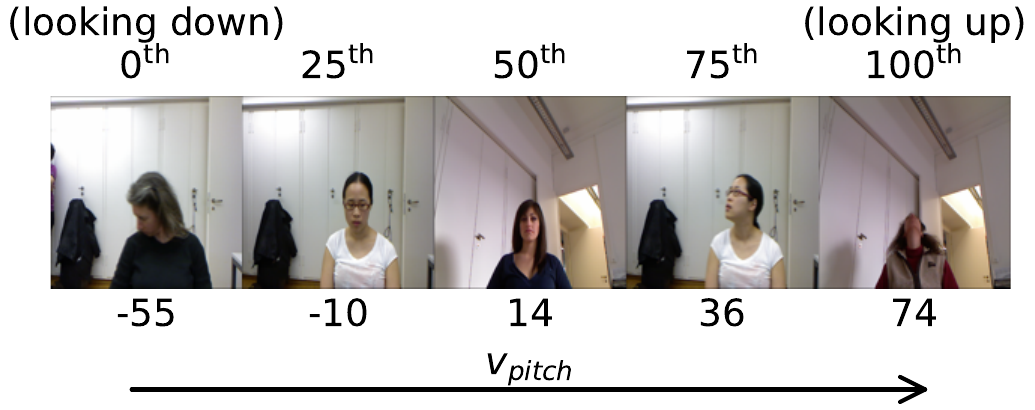}
  \end{subfigure}

  \vspace{1ex}

  \begin{subfigure}[t]{0.49\linewidth}
    \centering
    \includegraphics[width=\linewidth]{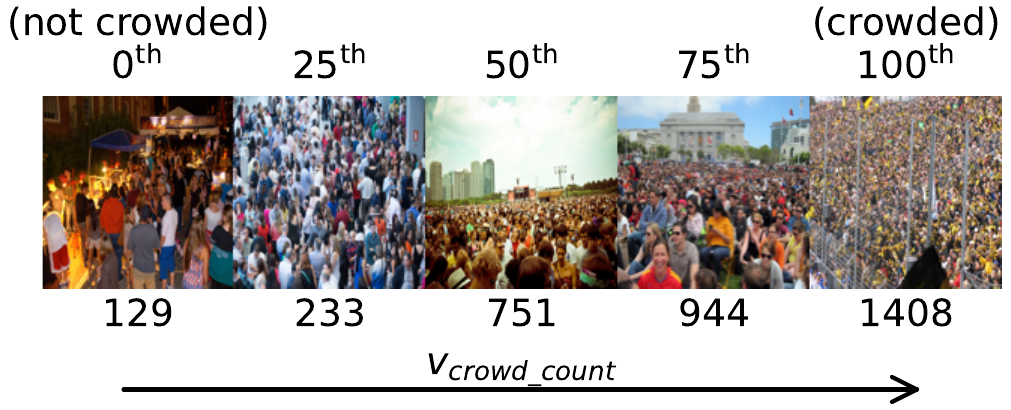}
  \end{subfigure}
  \hfill
  \begin{subfigure}[t]{0.49\linewidth}
    \centering
    \includegraphics[width=\linewidth]{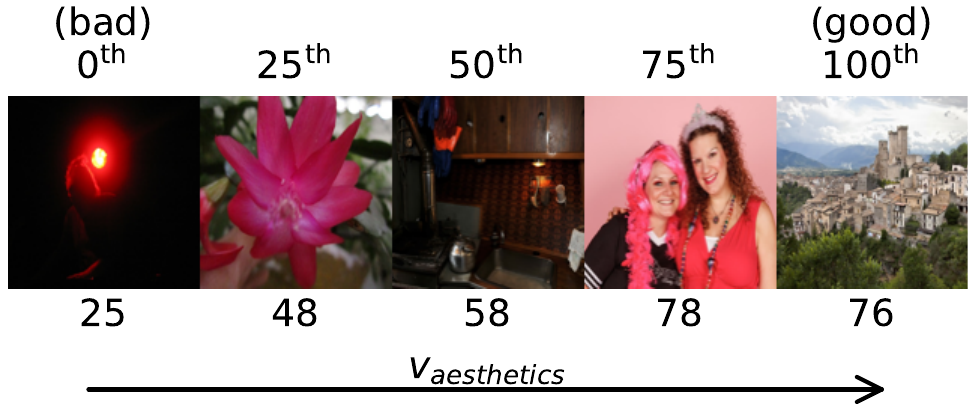}
  \end{subfigure}
  
  \caption{\textbf{Visualisation of rank axes.} We show $r^\text{th}$ percentile samples along the rank axes found using linear regression over CLIP-ViT-B/32 embeddings from each respective dataset.}
  
  \label{fig:dataset_visualization}
\end{figure}

\textbf{Setup.}
The SRCC for the linear regressor measures the rankability of the underlying vision encoder, while the baselines (no-encoder, nonlinear regression and finetuning) provide reference points. We report our main results in \cref{tab:all_metrics} and \cref{tab:model_wise_rankabilities}. In \cref{tab:all_metrics}, we average metrics across all architectures. Then in \cref{tab:model_wise_rankabilities}, we zoom into individual architectures while retaining only the rankability metric and averaging across all datasets that contain the same attribute. We also report qualitative results in \cref{fig:dataset_visualization}. Our observations vary across invididual attributes and architectures, but some patterns emerge.

\begin{table}[t]
  \centering \small
  \renewcommand{\arraystretch}{1.2}
  \setlength{\tabcolsep}{6pt}
  \caption{
  \textbf{Vision embeddings are generally rankable}.
  Spearman rank correlation $\rho$ between the true ranking of data samples and the predicted ranks. Higher is better. Results are averaged across all 7 architectures. \textbf{No-train}: Find $v_A$ on the embeddings of a randomly-initialised encoder. \textbf{Rankability}: How well does $v_A$ encode ordinal information? \textbf{Nonlinear}: Maximal non-linear ordinal information contained in embeddings. \textbf{Finetuned}: How learnable is the target attribute? For more information, see \cref{sec:defining_rankability}.
  }
  \vspace{0.5em}
       \begin{tabular}{l m{1.6cm} m{1.6cm} m{1.6cm} m{1.6cm}}

    \toprule
    \textbf{Attribute (Dataset)} 
    & \makecell{\textbf{No-train}\\lower bound} 
    & \makecell{\textbf{Rankability}\\main} 
    & \makecell{\textbf{Nonlinear}\\upper bound} 
    & \makecell{\textbf{Finetuned}\\upper bound} \\
    \origmidrule
    Age {\tiny (UTKFace)}              
    &\scorecell{myorange}{0.199}
    &\scorecell{mylgreen}{0.766}
    &\scorecell{mylgreen}{0.776}&
    \scorecell{mylgreen}{0.799}
    \\
    Age {\tiny (Adience) }
    &\scorecell{mylgreen}{0.266}
    &\scorecell{mydgreen}{0.861}
    &\scorecell{mydgreen}{0.878}
    &\scorecell{mydgreen}{0.910} 
    \\[0.5ex]
    \midrule
    Crowd {\tiny (UCF-QNRF)}      & \scorecell{myyellow}{0.220} & \scorecell{mydgreen}{0.843} & \scorecell{mydgreen}{0.854} & \scorecell{mydgreen}{0.886} \\
    Crowd Count {\tiny (ST-A)}& \scorecell{myorange}{0.120} & \scorecell{mylgreen}{0.734} & \scorecell{mylgreen}{0.749} & \scorecell{myyellow}{0.689} \\
    Crowd Count {\tiny (ST-B)}& \scorecell{myorange}{0.135} & \scorecell{mydgreen}{0.869} & \scorecell{mydgreen}{0.887} & \scorecell{mylgreen}{0.840} \\[0.5ex]
    \midrule
    Pitch {\tiny (Kinect)}              & \scorecell{myyellow}{0.405} & \scorecell{mylgreen}{0.803} & \scorecell{mylgreen}{0.811} & \scorecell{mydgreen}{0.975} \\
    Yaw {\tiny (Kinect)       }         & \scorecell{myred}{0.078}    & \scorecell{myorange}{0.434} & \scorecell{myyellow}{0.597} & \scorecell{mydgreen}{0.967} \\
    Roll {\tiny (Kinect)   }            & \scorecell{myorange}{0.151} & \scorecell{myorange}{0.218} & \scorecell{myyellow}{0.326} & \scorecell{mydgreen}{0.859} \\[0.5ex]
    \midrule
    Aesthetics {\tiny (AVA)   }         & \scorecell{myorange}{0.156} & \scorecell{myyellow}{0.653} & \scorecell{myyellow}{0.692} & \scorecell{myyellow}{0.693} \\
    Aesthetics {\tiny (KonIQ-10k) }      & \scorecell{myyellow}{0.435} & \scorecell{mylgreen}{0.761} & \scorecell{mylgreen}{0.793} & \scorecell{mydgreen}{0.901} \\[0.5ex]
    Recency {\tiny (HCI)    }           & \scorecell{myyellow}{0.324} & \scorecell{myyellow}{0.680} & \scorecell{myyellow}{0.688} & \scorecell{mylgreen}{0.722} \\
    \bottomrule
  \end{tabular}
  \label{tab:all_metrics}
\end{table}

\begin{table}[t]
  \centering
  \small
  \setlength{\tabcolsep}{7pt}
  \renewcommand{\arraystretch}{1}
  \caption{\textbf{Rankability across datasets and models}.
  Spearman's rank correlation $\rho$ between the true ranking of data samples and the predicted ranks. Higher is better. See Appendix \cref{tab:model_overview} for model details.}
  \vspace{6pt}
  \begin{tabular}{ >{\hspace{0pt}}m{9em} *{6}{>{\centering\arraybackslash}m{2.9em}} m{2em} }

    \toprule
     & \multicolumn{7}{c@{}}{\textbf{Model}} \\
  \cmidrule(l){2-8}
  \textbf{Attribute (Dataset)}& {RN50} & {ViTB32} & {CNX} & {DINO-B14} & {CLIP-RN50} & {CLIP-ViTB32} & {CLIP-CNX} \\
  \addlinespace[2pt]
  \origmidrule

    Age {\tiny (UTKFace)    }            & \heatcellB{0.633} & \heatcellB{0.739} & \heatcellB{0.772} & \heatcellB{0.770} & \heatcellB{0.820} & \heatcellB{0.810} & \heatcellB{0.820} \\
    Age {\tiny (Adience)  }             & \heatcellB{0.723} & \heatcellB{0.828} & \heatcellB{0.871} & \heatcellB{0.853} & \heatcellB{0.898} & \heatcellB{0.924} & \heatcellB{0.928} \\
    \midrule
    Crowd {\tiny (UCF-QNRF)     }       & \heatcellB{0.864} & \heatcellB{0.837} & \heatcellB{0.810} & \heatcellB{0.788} & \heatcellB{0.870} & \heatcellB{0.870} & \heatcellB{0.860} \\
    Crowd {\tiny (ST-A)   }             & \heatcellB{0.799} & \heatcellB{0.700} & \heatcellB{0.695} & \heatcellB{0.653} & \heatcellB{0.760} & \heatcellB{0.750} & \heatcellB{0.780} \\
    Crowd {\tiny (ST-B)    }             & \heatcellB{0.879} & \heatcellB{0.878} & \heatcellB{0.867} & \heatcellB{0.821} & \heatcellB{0.890} & \heatcellB{0.860} & \heatcellB{0.890} \\
    \midrule
    Pitch {\tiny (Kinect)     }        & \heatcellB{0.663} & \heatcellB{0.673} & \heatcellB{0.909} & \heatcellB{0.716} & \heatcellB{0.860} & \heatcellB{0.920} & \heatcellB{0.880} \\
    Yaw {\tiny (Kinect)     }           & \heatcellB{0.624} & \heatcellB{0.305} & \heatcellB{0.384} & \heatcellB{0.804} & \heatcellB{0.120} & \heatcellB{0.360} & \heatcellB{0.440} \\
    Roll {\tiny (Kinect)  }            & \heatcellB{0.352} & \heatcellB{0.196} & \heatcellB{0.298} & \heatcellB{0.512} & \heatcellB{0.090} & \heatcellB{0.020} & \heatcellB{0.060} \\
    \midrule
    Aesthetics {\tiny (AVA)  }             & \heatcellB{0.589} & \heatcellB{0.609} & \heatcellB{0.644} & \heatcellB{0.566} & \heatcellB{0.700} & \heatcellB{0.710} & \heatcellB{0.750} \\
    Aesthetics ({\tiny KonIQ-10k})         & \heatcellB{0.739} & \heatcellB{0.713} & \heatcellB{0.744} & \heatcellB{0.681} & \heatcellB{0.800} & \heatcellB{0.790} & \heatcellB{0.860} \\
    \midrule
    Recency {\tiny (HCI) }            & \heatcellB{0.600} & \heatcellB{0.592} & \heatcellB{0.631} & \heatcellB{0.571} & \heatcellB{0.780} & \heatcellB{0.770} & \heatcellB{0.820} \\
    \bottomrule
  \end{tabular}
  \label{tab:model_wise_rankabilities}
\end{table}

\textbf{Average rankability of vision embeddings is non-trivially high.}   
We observe in \cref{tab:all_metrics} that with respect to age on the Adience dataset, the average rankability of $0.861$ is much higher than the no-train lower bound of $0.266$, while the nonlinear and finetuned upper bounds ($0.878$ and $0.910$) are only slightly higher in comparison. In fact, for all attributes except for yaw and roll, the average rankability is closer to the two upper bounds than to the lower bound.

\textbf{CLIP embeddings are more rankable than non-CLIP embeddings.}
We observe in \cref{tab:model_wise_rankabilities} that on age, aesthetics, recency and pitch, the best CLIP encoder (\eg CLIP-ConvNeXt with an SRCC of $0.928$ on Adience) wins out against the best non-CLIP encoder (\eg vanilla ConvNeXt with an SRCC of $0.871$ on Adience). On crowd count, the best CLIP encoders are largely tied with the best non-CLIP encoders (\eg CLIP-RN50 at $0.870$ vs vanilla RN50 at $0.864$ for UCF-QNRF). On yaw and roll, DINO massively outperforms CLIP encoders (\eg $0.804$ vs $0.440$ for yaw). In conclusion, apart from isolated but interesting exceptions, CLIP encoders generally outperform or match non-CLIP encoders.

\textbf{Some attributes are better ranked than others.}
For example, the average SRCC over the two age datasets is $\sim 0.8$ (see \cref{tab:all_metrics}). Similar average SRCCs are observed for crowd count and pitch angle. Image aesthetics and recency are less well-ranked with average SRCCs between $0.65$ and $0.75$. Even within the same dataset (BIWI Kinect), yaw and roll angles with SRCCs of $0.434$ and $0.218$ are quite poorly ranked in comparison to pitch ($0.803$). We hypothesize that attribute-wise rankabilities are directly proportional to attribute-wise variety present in the training data. 

\vspace{-1.5em}

\textbf{Caveats.}
The current results are empirical, and our claims are based on the set of attributes considered in our study. Despite following choices established in the literature, we may sometimes not uncover the most optimal finetuned upper bounds. A broader study involving theoretical support for rankability, using more ordinal attributes and possibly stronger upper bounds would be a promising direction for future work.

\begin{tcolorbox}
\textbf{Takeaway from §3.} In general, visual embeddings are highly rankable compared to both the lower bound and upper bound baselines, although there exist variations across attributes (\eg most encoders struggle to rank based on yaw and roll angles) and encoders (\eg CLIP embeddings are, in general, more rankable than non-CLIP embeddings).
\end{tcolorbox}

\section{How to (efficiently) find the rank axis?}
\label{sec:finding-rank-axis}

In \cref{sec:are_embeddings_rankable}, we showed that visual embeddings are generally rankable. However, the rank axes $v_A$, determining the direction along which the continuous attribute $A$ is sorted accordingly (\cref{sec:defining_rankability}), were learned using a lot of training data with continuous-attribute annotations that are typically expensive to collect. We examine whether rank axes can also be discovered in more sample- and label-efficient manners.

We organise the section by first tackling an easier setting with abundant data and then more challenging settings with less data available. We begin with the few-shot setting with continuous attribute labels (\cref{sec:few_shot_cont}). Then, we examine the possibility to compute the rank axes with a few \quotes{extreme} points that require no cumbersome continuous attribute annotation (\cref{sec:few_shot_extreme}). 
For further practicality, we examine whether the obtained rank axes are resilient to domain shifts (\cref{sec:transfer}).
Finally, we discuss potential strategies to compute the rank axis in a zero-shot manner with text encoders in vision-language models (\cref{sec:zero_shot}).

\subsection{Few-shot learning with continuous attribute labels}
\label{sec:few_shot_cont}

In this section, we investigate the learnability of the rank axis $v_A$ for an attribute of interest $A$, when fewer samples with continuous attribute annotations are available. We report the results in \cref{fig:few_shot_rankability}. We choose the datasets UTKFace, Adience and AVA because of their relatively large sizes compared to the other datasets used in our experiments.

\paragraph{Only a fraction of training data is sufficient.} 
For age on the Adience dataset, only $\sim 1k$ training samples out of over $11k$ are sufficient for covering as much as $95\%$ of the gap between the SRCCs of the no-train baseline and full-dataset linear regression. Similarly, for image aesthetics on AVA (CLIP-ViT), only $\sim 16k$ (CLIP-ViT) or even $\sim 8k$ (CLIP-ConvNeXt) data points out of $\sim 230k$ are sufficient. It is evident that learning the rank axis often requires only a fraction of the original training dataset.

\subsection{Few-shot learning with extreme pairs}
\label{sec:few_shot_extreme}

We consider the practical scenario where a user wishes to sort their vector database using an arbitrary attribute $A$. They do not have access to continuous attribute annotations (as was assumed in \cref{sec:few_shot_cont}), but can readily obtain a few samples at the \quotes{extremes} of the desired rank axis. We test the effectiveness of this approach.

\textbf{Setup.} 
Given training and test splits $\mathcal{X}_{\text{train}}$ and $\mathcal{X}_{\text{test}}$, respectively, we sample sets of images $S_l$ and $S_u$ from the lower and upper extremes of $\mathcal{X}_{\text{train}}$, respectively. This simulates the scenario where a user may obtain such \quotes{extreme} images using a readily available source like a Web search engine. Next, we calculate the \quotes{lower extreme cluster} $x_l = \text{mean}(S_l)$ and \quotes{upper extreme cluster} $x_u = \text{mean}(S_u)$. Finally, the vector $v_{A} = \frac{x_u - x_l}{||x_u - x_l||}$ gives the rank axis (akin to a \textit{steering vector} \cite{rimsky2024steering, tigges2023lineara}).

\begin{figure}[H]
    \centering

    \begin{subfigure}[t]{0.32\textwidth}
        \centering
        \includegraphics[width=\linewidth]{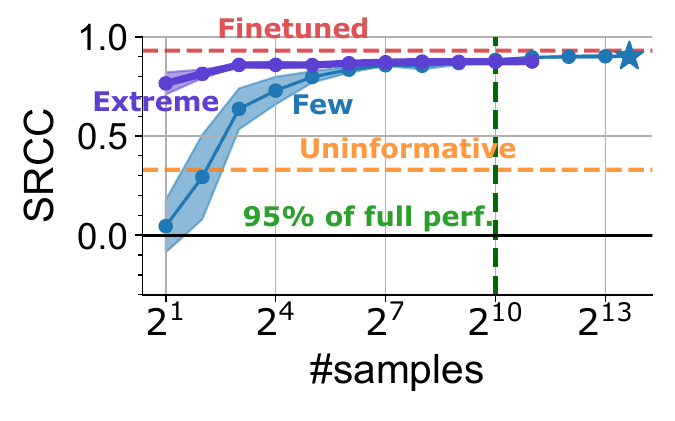}
        \caption{Adience (CLIP-ViT)}
    \end{subfigure}
    \hfill
    \begin{subfigure}[t]{0.32\textwidth}
        \centering
        \includegraphics[width=\linewidth]{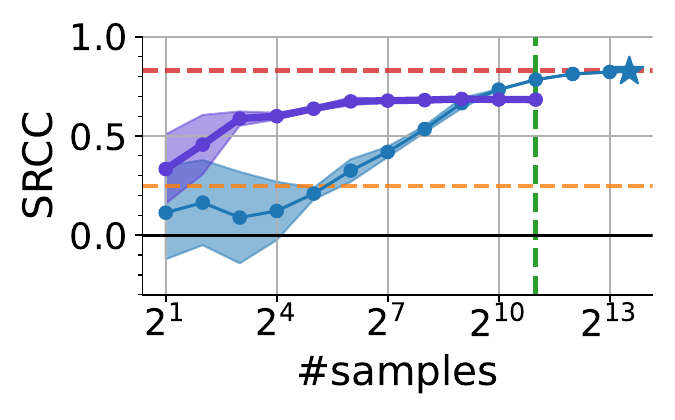}
        \caption{UTKFace (CLIP-ViT)}
    \end{subfigure}
    \hfill
    \begin{subfigure}[t]{0.32\textwidth}
        \centering
        \includegraphics[width=\linewidth]{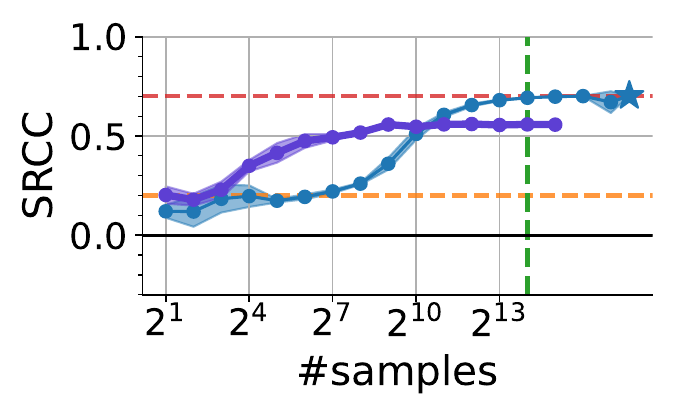}
        \caption{AVA (CLIP-ViT)}
    \end{subfigure}

    \begin{subfigure}[t]{0.32\textwidth}
        \centering
        \includegraphics[width=\linewidth]{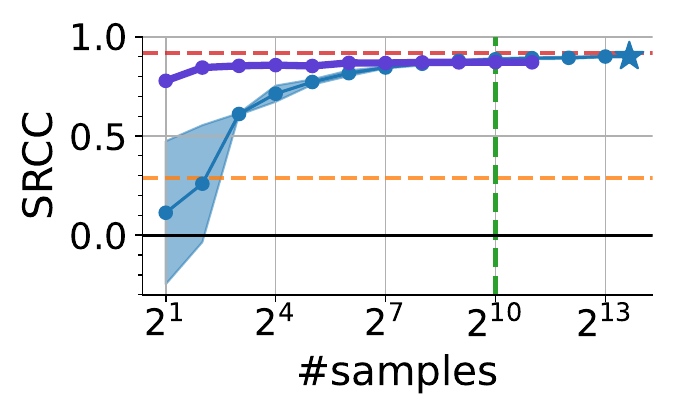}
        \caption{Adience (CLIP-ConvNeXt)}
    \end{subfigure}
    \hfill
    \begin{subfigure}[t]{0.32\textwidth}
        \centering
        \includegraphics[width=\linewidth]{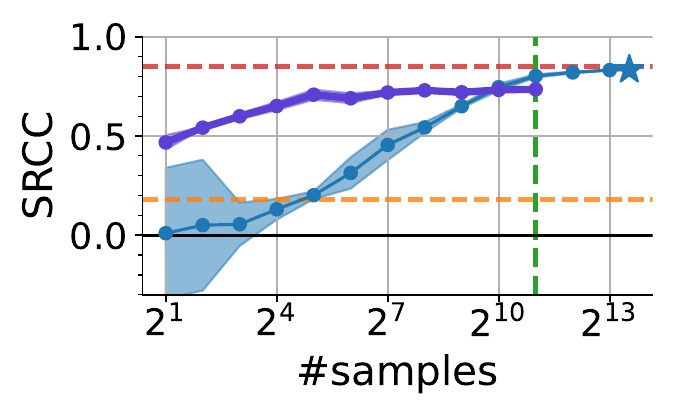}
        \caption{UTKFace (CLIP-ConvNeXt)}
    \end{subfigure}
    \hfill
    \begin{subfigure}[t]{0.32\textwidth}
        \centering
        \includegraphics[width=\linewidth]{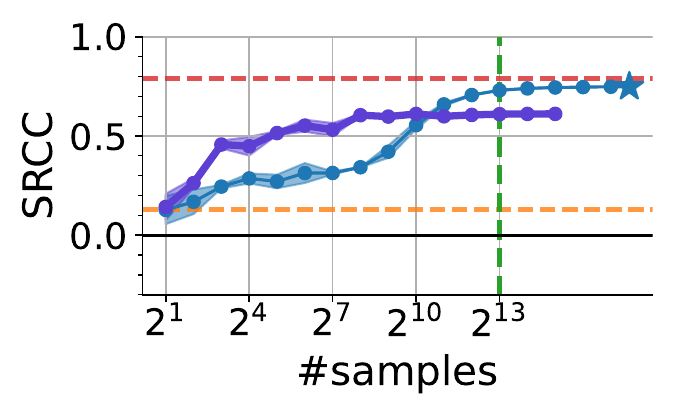}
        \caption{AVA (CLIP-ConvNeXt)}
    \end{subfigure}

    \caption{\textbf{Few-shot learning with continuous labels vs extreme samples without continuous labels: extreme samples win out in the small-train-set regime.} \quotes{Extreme} refers to training using samples from the extreme ends of the ranking axis, while "few" refers to few-shot learning on labeled samples. Full-dataset linear regression performance is given by the star-shaped marker.} 
    \label{fig:few_shot_rankability}
\end{figure}
\vspace{-1em}

\textbf{Observations.}
We report the results in \cref{fig:few_shot_rankability}, comparing with the few-shot scenario where GT continuous attribute labels are available. We observe that when the size of the training dataset is extremely small (upto $\sim 1k$), the extreme-pairs method performs better than or at par with few-shot training with continuous attribute labels. This effect can be seen most prominently in the Adience dataset where a rank axis obtained from just two extreme samples achieves an SRCC of $\sim 0.75$ on average, while the few-shot case results in an average SRCC close to $0$. As the training dataset continues to grow in size, few-shot training with labeled data catches up and finally surpasses the extreme-pairs method. It is nevertheless striking that extreme pairs perform so well at the lower ends of the few-shot setting.

\textbf{Takeaway.} If one has an extremely small number of samples ($1k$ or less), obtaining extreme samples from both ends of the rank axis (with no additional labeling) may consititute a better expenditure of resources than obtaining continuous attribute labels.

\subsection{Robustness of rank axis}
\label{sec:transfer}

We now consider the case where there is no access to samples from the target distribution. Assuming that a rank axes was previously learned using some source distribution, how transferable is it to other distributions? We investigate this using three attributes: age, crowd count, and image aesthetics.

\textbf{Setup}. In our main experiments, we use two age datasets (UTKFace, Adience), three crowd count datasets (UCF-QNRF, ShanghaiTech-A, and ShanghaiTech-B), and two aesthetics datasets (AVA, KonIQ-10k). Within each set of datasets containing the same attribute, we then test how well a rank axis learned from one dataset transfers to another, and vice versa. We employ SRCC on the target dataset as our transferability metric in \cref{tab:transfer}. Further, we also report cosine similarities between each pair of rank axes in \cref{tab:cosine}. As reference points, we also report inter-attribute observations (\eg transfrability from an age dataset to an aesthetics dataset). 

\textbf{Transfer is non-trivial and asymmetric.} For example, the rank axis learned from Adience has an SRCC of $0.680$ on UTKFace, which is significantly higher than the SRCCs of the same rank axis on other datasets. Given that the two datasets have different labeling systems (especially with Adience labels being much more coarse than those in UTKFace), this is not immediately intuitive, or trivial, and indicates the presence of some (albeit imperfect) \quotes{age} axis in the embedding space. At the same time, transfer in the opposite direction is significantly worse, \ie a rank axis trained on UTKFace achieves an SRCC of only $0.55$ on Adience. This observation most likely stems from the quality differences between the two datasets. 
    
\textbf{Rank directions are non-trivially correlated.} For example, the cosine similarity between age axes trained on UTKFace and Adience is $0.360$. While this similarity is much lower than $1.0$, it is still significant given the high dimensionality of the embedding space. This observation again indicates the presence of a \quotes{universal} age axis which was (albeit not perfectly) captured by regressors trained on both datasets. Notably, some unexpected correlations also emerge, \eg age (UTKFace) and aesthetics (KonIQ-10k) rank axes have a cosine similarity of $0.220$. This suggests the presence of unintended correlations in the training / test data.

\begin{table}[t]
  \centering
  \small
  \setlength{\tabcolsep}{3.5pt}
  \renewcommand{\arraystretch}{1.1}
  \caption{\textbf{Rank axis transferability}. Spearman rank correlation coefficients when a rank axis is \emph{trained} on dataset $i$ (rows) and
           \emph{evaluated} on dataset $j$ (columns).}
\vspace{.5em}

\begin{tabular}{l l *{7}{>{\centering\arraybackslash}m{4em}}}

  \toprule
    & & \multicolumn{7}{c}{\textbf{Evaluated on}}\\
    \cmidrule(lr){3-9}
    & &
      \makecell{Age\\{\tiny (UTKFace)}} &
      \makecell{Age\\{\tiny (Adience)}} &
      \makecell{Crowd\\{\tiny (UCF-QNRF)}} &
      \makecell{Crowd\\{\tiny (ST-A)}} &
      \makecell{Crowd\\{\tiny (ST-B)}} &
      \makecell{Aesthetics\\{\tiny (AVA)}} &
      \makecell{Aesthetics\\{\tiny (KonIQ10k)}} \\
    \midrule
    \multirow{7}{*}{\rotatebox{90}{\textbf{Trained on}}}\hspace{.3em}

      & \makecell[l]{Age {\tiny (UTKFace)}}         & \heatcellC{0.81} & \heatcellC{0.55} & \heatcellC{-0.12} & \heatcellC{-0.11} & \heatcellC{-0.15} & \heatcellC{-0.07} & \heatcellC{0.05} \\

      & \makecell[l]{Age {\tiny (Adience)}}         & \heatcellC{0.68} & \heatcellC{0.91} & \heatcellC{-0.13} & \heatcellC{-0.10} & \heatcellC{-0.21} & \heatcellC{0.01}  & \heatcellC{-0.08} \\

      & \makecell[l]{Crowd {\tiny (UCF-QNRF)}}      & \heatcellC{0.16} & \heatcellC{-0.39} & \heatcellC{0.87} & \heatcellC{0.82} & \heatcellC{0.66} & \heatcellC{0.01}  & \heatcellC{0.07} \\

      & \makecell[l]{Crowd {\tiny (ST-A)}}          & \heatcellC{0.05} & \heatcellC{-0.14} & \heatcellC{0.73} & \heatcellC{0.75} & \heatcellC{0.72} & \heatcellC{0.13}  & \heatcellC{0.07} \\

      & \makecell[l]{Crowd {\tiny (ST-B)}}          & \heatcellC{0.31} & \heatcellC{0.17}  & \heatcellC{0.41} & \heatcellC{0.39} & \heatcellC{0.86} & \heatcellC{0.11}  & \heatcellC{0.05} \\

      & \makecell[l]{Aesthetics {\tiny (AVA)}}      & \heatcellC{-0.12} & \heatcellC{-0.11} & \heatcellC{0.22} & \heatcellC{0.00} & \heatcellC{0.49} & \heatcellC{0.70}  & \heatcellC{0.45} \\

      & \makecell[l]{Aesthetics {\tiny (KonIQ10k)}} & \heatcellC{0.06} & \heatcellC{-0.08} & \heatcellC{0.03} & \heatcellC{0.18} & \heatcellC{0.05} & \heatcellC{0.28}  & \heatcellC{0.79} \\

  \bottomrule
\end{tabular}
\label{tab:transfer}
\end{table}

\vspace{-1em}

\begin{table}[ht]
  \centering 
  \footnotesize
  \setlength{\tabcolsep}{3.5pt}
  \renewcommand{\arraystretch}{1.1}
  \caption{\small \textbf{Cosine similarity of rank axes}. We compute geometric alignment of rank axes trained on dataset $i$ (rows) and dataset $j$ (columns).}
\vspace{.5em}
\begin{tabular}{l l *{7}{>{\centering\arraybackslash}m{4em}}}

  \toprule
    & & \multicolumn{7}{c}{\textbf{Dataset j}}\\
    \cmidrule(lr){3-9}
    & &
      \makecell{Age\\{\tiny (UTKFace)}} &
      \makecell{Age\\{\tiny (Adience)}} &
      \makecell{Crowd\\{\tiny (UCF-QNRF)}} &
      \makecell{Crowd\\{\tiny (ST-A)}} &
      \makecell{Crowd\\{\tiny (ST-B)}} &
      \makecell{Aesthetics\\{\tiny (AVA)}} &
      \makecell{Aesthetics\\{\tiny (KonIQ10k)}} \\
    \midrule
    \multirow{7}{*}{\rotatebox{90}{\textbf{Dataset i}}}\hspace{.3em}

      & \makecell[l]{Age {\tiny (UTKFace)}}         & \heatcellC{1.00} & \heatcellC{0.36} & \heatcellC{0.14} & \heatcellC{0.08} & \heatcellC{0.06} & \heatcellC{-0.04} & \heatcellC{0.22} \\

      & \makecell[l]{Age {\tiny (Adience)}}         & \heatcellC{0.36} & \heatcellC{1.00} & \heatcellC{0.02} & \heatcellC{0.04} & \heatcellC{0.03} & \heatcellC{0.03}  & \heatcellC{0.05} \\

      & \makecell[l]{Crowd {\tiny (UCF-QNRF)}}      & \heatcellC{0.14} & \heatcellC{0.02} & \heatcellC{1.00} & \heatcellC{0.54} & \heatcellC{0.26} & \heatcellC{-0.00} & \heatcellC{0.21} \\

      & \makecell[l]{Crowd {\tiny (ST-A)}}          & \heatcellC{0.08} & \heatcellC{0.04} & \heatcellC{0.54} & \heatcellC{1.00} & \heatcellC{0.31} & \heatcellC{0.01}  & \heatcellC{0.14} \\

      & \makecell[l]{Crowd {\tiny (ST-B)}}          & \heatcellC{0.06} & \heatcellC{0.03} & \heatcellC{0.26} & \heatcellC{0.31} & \heatcellC{1.00} & \heatcellC{0.08}  & \heatcellC{0.07} \\

      & \makecell[l]{Aesthetics {\tiny (AVA)}}      & \heatcellC{-0.04}& \heatcellC{0.03} & \heatcellC{-0.00}& \heatcellC{0.01} & \heatcellC{0.08} & \heatcellC{1.00}  & \heatcellC{0.29} \\

      & \makecell[l]{Aesthetics {\tiny (KonIQ10k)}} & \heatcellC{0.22} & \heatcellC{0.05} & \heatcellC{0.21} & \heatcellC{0.14} & \heatcellC{0.07} & \heatcellC{0.29}  & \heatcellC{1.00} \\

  \bottomrule
\end{tabular}
\label{tab:cosine}
\end{table}
\vspace{-0.5em}

\vspace{0.5em}
\subsection{Zero-shot setting}
\label{sec:zero_shot}

For VLMs, language is a potentially data-free approach to finding a rank axis. In principle, a text prompt could correspond to a rank axis in the embedding space. The SRCC of our linear regressors trained in \cref{sec:are_embeddings_rankable} then sets an upper bound to the SRCC of any rank axis recovered via prompting. In this section, we investigate the gap between this linear regression upper bound and zero-shot prompt search.

\paragraph{Setup.} We consider two zero-shot settings. In the \textbf{single-prompt} setting, the direction is defined by the embedding of a text prompt. In the \textbf{text-difference} setting, the direction is defined by the difference between the embeddings of two text prompts, each describing one extreme of the given attribute. We perform a prompt search over $500$ GPT-generated prompts for the single-prompt setting, and $100$ GPT-generated prompt pairs for the text-difference setting. We report our results in \cref{tab:zero_shot}.
\vspace{-1.5em}

\begin{wraptable}[14]{r}{0.55\linewidth} 
\vspace{-1.5em}
  \centering \small
  \renewcommand{\arraystretch}{1.2}
  \setlength{\tabcolsep}{6pt}
  \caption{\small 
  \textbf{Zero-shot results}.
  }
  \vspace{-0.7em}
  \begin{adjustbox}{width=\linewidth}
  \begin{tabular}{lccc}
    \toprule
    \textbf{Attribute (Dataset)} 
    & \makecell{\textbf{Zero-shot} \\ One prompt} 
    & \makecell{\textbf{Zero-shot}\\  Difference} 
    & \makecell{\textbf{Linear}\\upper bound} \\
    \midrule
    Age {\tiny (UTKFace)}           & \scorecell{mylgreen}{0.577} & \scorecell{mylgreen}{0.600} & \scorecell{mydgreen}{0.817} \\
    Age {\tiny (Adience)}           & \scorecell{mydgreen}{0.670} & \scorecell{mydgreen}{0.782} & \scorecell{mydgreen}{0.917} \\
    Crowd Count {\tiny (UCF-QNRF)}  & \scorecell{mylgreen}{0.523} & \scorecell{myorange}{0.315} & \scorecell{mydgreen}{0.867} \\
    Crowd Count {\tiny (ST-A)}      & \scorecell{mylgreen}{0.487} & \scorecell{myorange}{0.242} & \scorecell{mylgreen}{0.763} \\
    Crowd Count {\tiny (ST-B)}      & \scorecell{mylgreen}{0.590} & \scorecell{mylgreen}{0.535} & \scorecell{mydgreen}{0.880} \\
    Pitch {\tiny (Kinect)}          & \scorecell{mylgreen}{0.520} & \scorecell{mylgreen}{0.617} & \scorecell{mydgreen}{0.887} \\
    Yaw {\tiny (Kinect)}            & \scorecell{myorange}{0.117} & \scorecell{myorange}{0.060} & \scorecell{myyellow}{0.307} \\
    Roll {\tiny (Kinect)}           & \scorecell{myred}{-0.070}   & \scorecell{myred}{-0.010}   & \scorecell{myorange}{0.057} \\
    Aesthetics {\tiny (AVA)}        & \scorecell{myyellow}{0.367} & \scorecell{myyellow}{0.410} & \scorecell{mylgreen}{0.720} \\
    Aesthetics {\tiny (KonIQ-10k)}  & \scorecell{myorange}{0.103} & \scorecell{mylgreen}{0.547} & \scorecell{mydgreen}{0.817} \\
    Recency {\tiny (HCI)}           & \scorecell{myorange}{0.190} & \scorecell{myyellow}{0.449} & \scorecell{mydgreen}{0.790} \\
    \bottomrule
  \end{tabular}
  \end{adjustbox}
  \label{tab:zero_shot}
\end{wraptable}

\vspace{1.5em}

\paragraph{Observations and takeaway.} 
Overall, zero-shot methods are suboptimal. For instance, even the best-performing zero-shot approach, our difference-based method on Adience, achieves $\rho = 0.782$ vs. the corresponding linear model at $\rho = 0.917$. The gap is more pronounced for some attributes (\eg, $\rho = 0.449$ vs. $\rho = 0.790$ for image recency). While more optimal prompts may have been overlooked during the search, this also reflects a realistic setting where one exhausts all intuitive prompt choices. Currently, it is evident that language-based prompting lags considerably behind linear regression using image data, although the text-difference method improves upon vanilla prompting.

\begin{tcolorbox}
\textbf{Takeaway from §4.} One may often efficiently discover the (almost) optimal rank axis using only a fraction of the total labeled data. Learned rank axes are quite transferable across different datasets, suggesting the presence of \quotes{universal} rank axes. 
\end{tcolorbox}

\section{Conclusion and future work}

In this work, we investigate visual encoders for the presence of ordinal information. Extensive experiments reveal not only that such information is present, as indicated by the high Spearman rank correlation of nonlinear regressors, but also that most of the available ordinality is \textbf{linearly} encoded, as indicated by the small gap between the performances of linear and MLP regressors. The embedding space indeed possesses a \quotes{rankable} structure. This is unexpected and practically useful.

These findings provoke further questions. Most notably, the \textit{linearity} of ordinal information suggests that one could potentially also characterise embeddings as interpretable collections of latent ordinal subspaces. This would involve discovering a far bigger set of ordinal attributes: we leave this exciting direction to future work.

\paragraph{Acknowledgements.} This work was supported by the German Federal Ministry of Education and Research (BMBF): T{\"u}bingen AI Center, FKZ: 01IS18039A. The authors thank the International Max Planck Research School for Intelligent Systems (IMPRS-IS) for supporting Ankit Sonthalia and Arnas Uselis. The authors are also grateful to Wei-Hsiang Yu, the first author of \cite{yu2025rankingaware}, for helpful insights.

\printbibliography
\appendix

\clearpage
\renewcommand{\thesection}{\Alph{section}}
\renewcommand{\thesubsection}{\thesection.\arabic{subsection}}
\renewcommand{\thesubsubsection}{\thesubsection.\arabic{subsubsection}}
\renewcommand{\thetable}{\thesection\arabic{table}}

\section{Further Experimental Details}

Here, we list further details of the architectures considered in our study.

\begin{table}[H]
  \centering
  \caption{\textbf{Summary of vision encoders used in our study.} “Acronym” refers to how each model is denoted in our main results tables.}
  \vspace{6pt}
  \resizebox{\textwidth}{!}{%
  \begin{tabular}{@{}l l c c c l c@{}}
    \toprule
    \textbf{Acronym} & \textbf{Architecture} & \textbf{\#Dims} & \textbf{Year} & \textbf{\#Params} & \textbf{Type} & \textbf{Input Size} \\
    \midrule
    RN50             & ResNet50             & 2048 & 2015 & 25.6M   & ConvNet     & 224×224 \\
    ViTB32           & ViT-B/32             & 768  & 2020 & 88.2M   & Transformer & 224×224 \\
    CNX              & ConvNeXtV2           & 1536 & 2023 & 198M    & ConvNet     & 224×224 \\
    DINO-B14         & DINOv2 (ViT-B/14)    & 768  & 2023 & 86.6M   & Transformer & 518×518 \\
    CLIP-RN50        & CLIP ResNet50        & 1024 & 2021 & 38.3M   & ConvNet     & 224×224 \\
    CLIP-ViTB32      & CLIP ViT-B/32        & 512  & 2021 & 87.8M   & Transformer & 224×224 \\
    CLIP-CNX         & CLIP ConvNeXtV2      & 768  & 2023 & 199.8M  & ConvNet     & 320×320 \\
    \bottomrule
  \end{tabular}
  }
  \label{tab:model_overview}
\end{table}

\section{Further details on datasets and dataset-specific results}
\label{supp:datasets}

In the main paper, we present results for each dataset aggregated over all models (Table 2 in the main paper) and rankabilities (Spearman $\rho$) for each model-dataset pair (Table 3 in the main paper). While the main results convey the primary evidence towards our claim (vision embeddings are rankable), we also report more detailed results on each dataset in this section. Please also refer to Table 1 in the main paper for a condensed overview of all datasets considered.

\subsection{Age}

\textbf{UTKFace}, introduced in \cite{zhang2017age}, is a dataset of face images with age labels ranging from 0 to 116. Following \cite{yu2025rankingaware, kuprashevich2023mivolo}, we use a smaller subset with ages ranging between 21 and 60. The dataset was downloaded from the official website (\href{https://susanqq.github.io/UTKFace/}{https://susanqq.github.io/UTKFace/}). We report results in \cref{tab:utkface}. 

\begin{table}[H]
  \centering
  \small
  \setlength{\tabcolsep}{6pt}
  \renewcommand{\arraystretch}{1.2}
  \caption{
  \textbf{UTKFace (Age)}. Spearman’s rank correlation $\rho$ across evaluation strategies. 
  \textbf{No-train}: linear probe on untrained encoder. 
  \textbf{Rankability}: linear probe on pretrained encoder. 
  \textbf{Nonlinear}: MLP on frozen encoder. 
  \textbf{Finetuned}: encoder + head trained end-to-end.
  Higher is better.}
  \vspace{6pt}
  \begin{tabular}{lcccc}
    \toprule
    \textbf{Model} 
    & \makecell{\textbf{No-train}\\lower bound} 
    & \makecell{\textbf{Rankability}\\main} 
    & \makecell{\textbf{Nonlinear}\\upper bound} 
    & \makecell{\textbf{Finetuned}\\upper bound} \\
    \midrule
    ResNet-50                    
    & \heatcellB{0.212} 
    & \heatcellB{0.633} 
    & \heatcellB{0.636} 
    & \heatcellB{0.762} \\
    
    ViT-B/32                     
    & \heatcellB{0.283} 
    & \heatcellB{0.739} 
    & \heatcellB{0.749} 
    & \heatcellB{0.737} \\
    
    ConvNeXtV2-L                 
    & \heatcellB{0.283} 
    & \heatcellB{0.772} 
    & \heatcellB{0.776} 
    & \heatcellB{0.815} \\
    
    DINOv2 ViT-B/14               
    & \heatcellB{0.054} 
    & \heatcellB{0.770} 
    & \heatcellB{0.770} 
    & \heatcellB{0.812} \\
    
    OpenAI CLIP ResNet-50                 
    & \heatcellB{0.130} 
    & \heatcellB{0.820} 
    & \heatcellB{0.830} 
    & \heatcellB{0.790} \\
    
    OpenAI CLIP ViT-B/32                  
    & \heatcellB{0.250} 
    & \heatcellB{0.810} 
    & \heatcellB{0.830} 
    & \heatcellB{0.830} \\
    
    OpenCLIP ConvNeXt-L (D, 320px)        
    & \heatcellB{0.180} 
    & \heatcellB{0.820} 
    & \heatcellB{0.840} 
    & \heatcellB{0.850} \\

        \midrule
    \textbf{Mean}
    & \heatcellB{0.199} 
    & \heatcellB{0.766} 
    & \heatcellB{0.776} 
    & \heatcellB{0.799} \\

    \bottomrule
  \end{tabular}
  \label{tab:utkface}
\end{table}

\textbf{Adience}, introduced in \cite{eidinger2014age}, is another age dataset. Unlike UTKFace, it contains coarse labels (8 age groups instead of exact ages). We use the \quotes{aligned} version of the images and five-fold cross-validation as in \cite{yu2025rankingaware}. Results can be found in \cref{tab:adience}.

\begin{table}[H]
  \centering
  \small
  \setlength{\tabcolsep}{6pt}
  \renewcommand{\arraystretch}{1.2}
  \caption{
  \textbf{Adience (Age)}. Spearman’s rank correlation $\rho$ across evaluation strategies.  }
  \vspace{6pt}
  \begin{tabular}{lcccc}
    \toprule
    \textbf{Model} 
    & \makecell{\textbf{No-train}\\lower bound} 
    & \makecell{\textbf{Rankability}\\main} 
    & \makecell{\textbf{Nonlinear}\\upper bound} 
    & \makecell{\textbf{Finetuned}\\upper bound} \\
    \midrule
    ResNet-50                    
    & \heatcellB{0.328} 
    & \heatcellB{0.723} 
    & \heatcellB{0.759} 
    & \heatcellB{0.894} \\
    
    ViT-B/32                        
    & \heatcellB{0.310} 
    & \heatcellB{0.828} 
    & \heatcellB{0.860} 
    & \heatcellB{0.892} \\
    
    ConvNeXtV2-L                   
    & \heatcellB{0.522} 
    & \heatcellB{0.871} 
    & \heatcellB{0.885} 
    & \heatcellB{0.910} \\
    
    DINOv2 ViT-B/14                
    & \heatcellB{0.120} 
    & \heatcellB{0.853} 
    & \heatcellB{0.877} 
    & \heatcellB{0.914} \\
    
    OpenAI CLIP ResNet-50                 
    & \heatcellB{0.070} 
    & \heatcellB{0.898} 
    & \heatcellB{0.914} 
    & \heatcellB{0.894} \\
    
    OpenAI CLIP ViT-B/32                  
    & \heatcellB{0.292} 
    & \heatcellB{0.924} 
    & \heatcellB{0.922} 
    & \heatcellB{0.928} \\
    
    OpenCLIP ConvNeXt-L (D, 320px)        
    & \heatcellB{0.220} 
    & \heatcellB{0.928} 
    & \heatcellB{0.932} 
    & \heatcellB{0.938} \\

   \midrule
    \textbf{Mean}
    & \heatcellB{0.266} 
    & \heatcellB{0.861} 
    & \heatcellB{0.878} 
    & \heatcellB{0.910} \\

    \bottomrule
  \end{tabular}
  \label{tab:adience}
\end{table}

\subsection{Crowd count}

\textbf{UCF-QNRF}, introduced in \cite{idrees2018composition}, is a large crowd counting dataset containing images from diverse parts of the world. We use the official download link at \href{https://www.crcv.ucf.edu/data/ucf-qnrf/}{https://www.crcv.ucf.edu/data/ucf-qnrf/} and the official train-test splits. Results can be found in \cref{tab:ucfqnrf_modelwise}.

\begin{table}[H]
  \centering
  \small
  \setlength{\tabcolsep}{6pt}
  \renewcommand{\arraystretch}{1.2}
  \caption{
  \textbf{UCF-QNRF (Crowd Count)}. Spearman’s rank correlation $\rho$ across evaluation strategies.}
  \vspace{6pt}
  \begin{tabular}{lcccc}
    \toprule
    \textbf{Model} 
    & \makecell{\textbf{No-train}\\lower bound} 
    & \makecell{\textbf{Rankability}\\main} 
    & \makecell{\textbf{Nonlinear}\\upper bound} 
    & \makecell{\textbf{Finetuned}\\upper bound} \\
    \midrule
    ResNet-50                    
    & \heatcellB{0.466} 
    & \heatcellB{0.864} 
    & \heatcellB{0.870} 
    & \heatcellB{0.826} \\
    
    ViT-B/32                     
    & \heatcellB{0.288} 
    & \heatcellB{0.837} 
    & \heatcellB{0.840} 
    & \heatcellB{0.794} \\
    
    ConvNeXtV2-L                
    & \heatcellB{-0.054} 
    & \heatcellB{0.810} 
    & \heatcellB{0.816} 
    & \heatcellB{0.938} \\
    
    DINOv2 ViT-B/14              
    & \heatcellB{0.219} 
    & \heatcellB{0.788} 
    & \heatcellB{0.842} 
    & \heatcellB{0.961} \\
    
    OpenAI CLIP ResNet-50                 
    & \heatcellB{0.240} 
    & \heatcellB{0.870} 
    & \heatcellB{0.880} 
    & \heatcellB{0.820} \\
    
    OpenAI CLIP ViT-B/32                  
    & \heatcellB{0.280} 
    & \heatcellB{0.870} 
    & \heatcellB{0.870} 
    & \heatcellB{0.900} \\
    
    OpenCLIP ConvNeXt-L (D, 320px)        
    & \heatcellB{0.100} 
    & \heatcellB{0.860} 
    & \heatcellB{0.860} 
    & \heatcellB{0.960} \\

    \midrule
    \textbf{Mean}
    & \heatcellB{0.220} 
    & \heatcellB{0.843} 
    & \heatcellB{0.854} 
    & \heatcellB{0.886} \\

    \bottomrule
  \end{tabular}
  \label{tab:ucfqnrf_modelwise}
\end{table}

\textbf{ShanghaiTech}, introduced in \cite{zhang2016singleimage}, is another crowd counting dataset consisting of two parts: A and B. While part A was crawled from the Internet and features larger crowds in general, part B was taken from metropolitan areas of Shanghai and features much smaller crowds. We use the DropBox link available at \href{https://github.com/desenzhou/ShanghaiTechDataset}{https://github.com/desenzhou/ShanghaiTechDataset} and official train-test splits for both parts. Results can be found in \cref{tab:shanghaitech_a_modelwise} and \cref{tab:shanghaitech_b_modelwise}.

\begin{table}[H]
  \centering
  \small
  \setlength{\tabcolsep}{6pt}
  \renewcommand{\arraystretch}{1.2}
  \caption{
  \textbf{ShanghaiTech-A (Crowd Count)}. Spearman’s rank correlation $\rho$ across evaluation strategies.}
  \vspace{6pt}
  \begin{tabular}{lcccc}
    \toprule
    \textbf{Model} 
    & \makecell{\textbf{No-train}\\lower bound} 
    & \makecell{\textbf{Rankability}\\main} 
    & \makecell{\textbf{Nonlinear}\\upper bound} 
    & \makecell{\textbf{Finetuned}\\upper bound} \\
    \midrule
    ResNet-50                      
    & \heatcellB{0.359} 
    & \heatcellB{0.799} 
    & \heatcellB{0.802} 
    & \heatcellB{0.529} \\
    
    ViT-B/32                       
    & \heatcellB{0.148} 
    & \heatcellB{0.700} 
    & \heatcellB{0.623} 
    & \heatcellB{0.558} \\
    
    ConvNeXtV2-L                  
    & \heatcellB{0.061} 
    & \heatcellB{0.695} 
    & \heatcellB{0.753} 
    & \heatcellB{0.834} \\
    
    DINOv2 ViT-B/14                
    & \heatcellB{-0.017} 
    & \heatcellB{0.653} 
    & \heatcellB{0.722} 
    & \heatcellB{0.786} \\
    
    OpenAI CLIP ResNet-50                 
    & \heatcellB{0.200} 
    & \heatcellB{0.760} 
    & \heatcellB{0.770} 
    & \heatcellB{0.510} \\
    
    OpenAI CLIP ViT-B/32                  
    & \heatcellB{0.010} 
    & \heatcellB{0.750} 
    & \heatcellB{0.770} 
    & \heatcellB{0.700} \\
    
    OpenCLIP ConvNeXt-L (D, 320px)        
    & \heatcellB{0.080} 
    & \heatcellB{0.780} 
    & \heatcellB{0.800} 
    & \heatcellB{0.910} \\

        \midrule
    \textbf{Mean}
    & \heatcellB{0.120} 
    & \heatcellB{0.734} 
    & \heatcellB{0.749} 
    & \heatcellB{0.689} \\

    \bottomrule
  \end{tabular}
  \label{tab:shanghaitech_a_modelwise}
\end{table}

\begin{table}[H]
  \centering
  \small
  \setlength{\tabcolsep}{6pt}
  \renewcommand{\arraystretch}{1.2}
  \caption{
  \textbf{ShanghaiTech-B (Crowd Count)}. Spearman’s rank correlation $\rho$ across evaluation strategies.}
  \vspace{6pt}
  \begin{tabular}{lcccc}
    \toprule
    \textbf{Model} 
    & \makecell{\textbf{No-train}\\lower bound} 
    & \makecell{\textbf{Rankability}\\main} 
    & \makecell{\textbf{Nonlinear}\\upper bound} 
    & \makecell{\textbf{Finetuned}\\upper bound} \\
    \midrule
    ResNet-50                      
    & \heatcellB{0.280} 
    & \heatcellB{0.879} 
    & \heatcellB{0.906} 
    & \heatcellB{0.672} \\
    
    ViT-B/32                       
    & \heatcellB{0.225} 
    & \heatcellB{0.878} 
    & \heatcellB{0.889} 
    & \heatcellB{0.710} \\
    
    ConvNeXtV2-L                  
    & \heatcellB{0.070} 
    & \heatcellB{0.867} 
    & \heatcellB{0.876} 
    & \heatcellB{0.955} \\
    
    DINOv2 ViT-B/14                
    & \heatcellB{0.020} 
    & \heatcellB{0.821} 
    & \heatcellB{0.869} 
    & \heatcellB{0.972} \\
    
    OpenAI CLIP ResNet-50                 
    & \heatcellB{0.270} 
    & \heatcellB{0.890} 
    & \heatcellB{0.900} 
    & \heatcellB{0.690} \\
    
    OpenAI CLIP ViT-B/32                  
    & \heatcellB{0.040} 
    & \heatcellB{0.860} 
    & \heatcellB{0.860} 
    & \heatcellB{0.900} \\
    
    OpenCLIP ConvNeXt-L (D, 320px)        
    & \heatcellB{0.040} 
    & \heatcellB{0.890} 
    & \heatcellB{0.910} 
    & \heatcellB{0.980} \\

        \midrule
    \textbf{Mean}
    & \heatcellB{0.135} 
    & \heatcellB{0.869} 
    & \heatcellB{0.887} 
    & \heatcellB{0.840} \\

    \bottomrule
  \end{tabular}
  \label{tab:shanghaitech_b_modelwise}
\end{table}

\subsection{Headpose (Euler angles)}

The \textbf{BIWI Kinect} dataset, introduced in \cite{fanelli2011real}, is a collection of 24 different videos wherein the subject of the video sits about a meter away from a Kinect (\href{https://en.wikipedia.org/wiki/Kinect}{https://en.wikipedia.org/wiki/Kinect}) sensor and rotates their head to span the entire range of possible head-pose angles pitch (rotation about the x-axis), yaw (rotation about the y-axis) and roll (rotation about the z-axis). As there exists no official split that we know of, we randomly hold out 6 sequences for testing. Results can be found in \cref{tab:kinect_pitch_modelwise} (pitch), \cref{tab:kinect_yaw_modelwise} (yaw) and \cref{tab:kinect_roll_modelwise} (roll).

\begin{table}[H]
  \centering
  \small
  \setlength{\tabcolsep}{6pt}
  \renewcommand{\arraystretch}{1.2}
  \caption{
  \textbf{Kinect (Pitch)}. Spearman’s rank correlation $\rho$ across evaluation strategies.}
  \vspace{6pt}
  \begin{tabular}{lcccc}
    \toprule
    \textbf{Model} 
    & \makecell{\textbf{No-train}\\lower bound} 
    & \makecell{\textbf{Rankability}\\main} 
    & \makecell{\textbf{Nonlinear}\\upper bound} 
    & \makecell{\textbf{Finetuned}\\upper bound} \\
    \midrule
    ResNet-50                      
    & \heatcellB{0.359} 
    & \heatcellB{0.663} 
    & \heatcellB{0.615} 
    & \heatcellB{0.973} \\
    
    ViT-B/32                       
    & \heatcellB{0.401} 
    & \heatcellB{0.673} 
    & \heatcellB{0.505} 
    & \heatcellB{0.951} \\
    
    ConvNeXtV2-L                  
    & \heatcellB{0.548} 
    & \heatcellB{0.909} 
    & \heatcellB{0.882} 
    & \heatcellB{0.984} \\
    
    DINOv2 ViT-B/14                
    & \heatcellB{0.231} 
    & \heatcellB{0.716} 
    & \heatcellB{0.986} 
    & \heatcellB{0.979} \\
    
    OpenAI CLIP ResNet-50                 
    & \heatcellB{0.450} 
    & \heatcellB{0.860} 
    & \heatcellB{0.870} 
    & \heatcellB{0.970} \\
    
    OpenAI CLIP ViT-B/32                  
    & \heatcellB{0.400} 
    & \heatcellB{0.920} 
    & \heatcellB{0.940} 
    & \heatcellB{0.980} \\
    
    OpenCLIP ConvNeXt-L (D, 320px)        
    & \heatcellB{0.450} 
    & \heatcellB{0.880} 
    & \heatcellB{0.880} 
    & \heatcellB{0.990} \\

        \midrule
    \textbf{Mean}
    & \heatcellB{0.405} 
    & \heatcellB{0.803} 
    & \heatcellB{0.811} 
    & \heatcellB{0.975} \\

    \bottomrule
  \end{tabular}
  \label{tab:kinect_pitch_modelwise}
\end{table}

\begin{table}[H]
  \centering
  \small
  \setlength{\tabcolsep}{6pt}
  \renewcommand{\arraystretch}{1.2}
  \caption{
  \textbf{Kinect (Yaw)}. Spearman’s rank correlation $\rho$ across evaluation strategies.}
  \vspace{6pt}
  \begin{tabular}{lcccc}
    \toprule
    \textbf{Model} 
    & \makecell{\textbf{No-train}\\lower bound} 
    & \makecell{\textbf{Rankability}\\main} 
    & \makecell{\textbf{Nonlinear}\\upper bound} 
    & \makecell{\textbf{Finetuned}\\upper bound} \\
    \midrule
    ResNet-50                      
    & \heatcellB{0.160} 
    & \heatcellB{0.624} 
    & \heatcellB{0.726} 
    & \heatcellB{0.990} \\
    
    ViT-B/32                       
    & \heatcellB{0.209} 
    & \heatcellB{0.305} 
    & \heatcellB{0.305} 
    & \heatcellB{0.838} \\
    
    ConvNeXtV2-L                  
    & \heatcellB{0.113} 
    & \heatcellB{0.384} 
    & \heatcellB{0.716} 
    & \heatcellB{0.989} \\
    
    DINOv2 ViT-B/14                
    & \heatcellB{-0.046} 
    & \heatcellB{0.804} 
    & \heatcellB{0.871} 
    & \heatcellB{0.994} \\
    
    OpenAI CLIP ResNet-50                 
    & \heatcellB{-0.060} 
    & \heatcellB{0.120} 
    & \heatcellB{0.530} 
    & \heatcellB{0.980} \\
    
    OpenAI CLIP ViT-B/32                  
    & \heatcellB{0.160} 
    & \heatcellB{0.360} 
    & \heatcellB{0.330} 
    & \heatcellB{0.990} \\
    
    OpenCLIP ConvNeXt-L (D, 320px)        
    & \heatcellB{0.010} 
    & \heatcellB{0.440} 
    & \heatcellB{0.700} 
    & \heatcellB{0.990} \\

        \midrule
    \textbf{Mean}
    & \heatcellB{0.078} 
    & \heatcellB{0.434} 
    & \heatcellB{0.597} 
    & \heatcellB{0.967} \\

    \bottomrule
  \end{tabular}
  \label{tab:kinect_yaw_modelwise}
\end{table}

\begin{table}[H]
  \centering
  \small
  \setlength{\tabcolsep}{6pt}
  \renewcommand{\arraystretch}{1.2}
  \caption{
  \textbf{Kinect (Roll)}. Spearman’s rank correlation $\rho$ across evaluation strategies.}
  \vspace{6pt}
  \begin{tabular}{lcccc}
    \toprule
    \textbf{Model} 
    & \makecell{\textbf{No-train}\\lower bound} 
    & \makecell{\textbf{Rankability}\\main} 
    & \makecell{\textbf{Nonlinear}\\upper bound} 
    & \makecell{\textbf{Finetuned}\\upper bound} \\
    \midrule
    ResNet-50                      
    & \heatcellB{0.202} 
    & \heatcellB{0.352} 
    & \heatcellB{0.430} 
    & \heatcellB{0.930} \\
    
    ViT-B/32                       
    & \heatcellB{0.098} 
    & \heatcellB{0.196} 
    & \heatcellB{0.375} 
    & \heatcellB{0.477} \\
    
    ConvNeXtV2-L                  
    & \heatcellB{0.550} 
    & \heatcellB{0.298} 
    & \heatcellB{0.368} 
    & \heatcellB{0.963} \\
    
    DINOv2 ViT-B/14                
    & \heatcellB{0.256} 
    & \heatcellB{0.512} 
    & \heatcellB{0.551} 
    & \heatcellB{0.912} \\
    
    OpenAI CLIP ResNet-50                 
    & \heatcellB{0.140} 
    & \heatcellB{0.090} 
    & \heatcellB{0.300} 
    & \heatcellB{0.920} \\
    
    OpenAI CLIP ViT-B/32                  
    & \heatcellB{-0.060} 
    & \heatcellB{0.020} 
    & \heatcellB{0.170} 
    & \heatcellB{0.850} \\
    
    OpenCLIP ConvNeXt-L (D, 320px)        
    & \heatcellB{-0.130} 
    & \heatcellB{0.060} 
    & \heatcellB{0.090} 
    & \heatcellB{0.960} \\

        \midrule
    \textbf{Mean}
    & \heatcellB{0.151} 
    & \heatcellB{0.218} 
    & \heatcellB{0.326} 
    & \heatcellB{0.859} \\

    \bottomrule
  \end{tabular}
  \label{tab:kinect_roll_modelwise}
\end{table}

\subsection{Aesthetics (Mean Opinion Score)}

The \textbf{Aesthetics Visual Analysis (AVA)} dataset, introduced in \cite{murray2012ava}, is a large-scale dataset including aesthetic preference scores provided by human annotators. Each image is labeled by multiple annotators, each assigning a score in the range 1-10. The mean opinion score (MOS) of the image is then computed as a weighted average over the ratings where the weight of a rating is provided by its frequency. We use the split provided by \cite{yu2025rankingaware} in their official repository (\href{https://github.com/uynaes/RankingAwareCLIP/tree/main/examples}{https://github.com/uynaes/RankingAwareCLIP/tree/main/examples}). Results are reported in \cref{tab:ava_modelwise}.

\begin{table}[H]
  \centering
  \small
  \setlength{\tabcolsep}{6pt}
  \renewcommand{\arraystretch}{1.2}
  \caption{
  \textbf{AVA (Image Aesthetics)}. Spearman’s rank correlation $\rho$ across evaluation strategies.}
  \vspace{6pt}
  \begin{tabular}{lcccc}
    \toprule
    \textbf{Model} 
    & \makecell{\textbf{No-train}\\lower bound} 
    & \makecell{\textbf{Rankability}\\main} 
    & \makecell{\textbf{Nonlinear}\\upper bound} 
    & \makecell{\textbf{Finetuned}\\upper bound} \\
    \midrule
    ResNet-50                      
    & \heatcellB{0.237} 
    & \heatcellB{0.589} 
    & \heatcellB{0.628} 
    & \heatcellB{0.672} \\
    
    ViT-B/32                       
    & \heatcellB{0.157} 
    & \heatcellB{0.609} 
    & \heatcellB{0.666} 
    & \heatcellB{0.672} \\
    
    ConvNeXtV2-L                  
    & \heatcellB{0.158} 
    & \heatcellB{0.644} 
    & \heatcellB{0.685} 
    & \heatcellB{0.728} \\
    
    DINOv2 ViT-B/14                
    & \heatcellB{0.057} 
    & \heatcellB{0.566} 
    & \heatcellB{0.648} 
    & \heatcellB{0.590} \\
    
    OpenAI CLIP ResNet-50                 
    & \heatcellB{0.150} 
    & \heatcellB{0.700} 
    & \heatcellB{0.710} 
    & \heatcellB{0.700} \\
    
    OpenAI CLIP ViT-B/32                  
    & \heatcellB{0.200} 
    & \heatcellB{0.710} 
    & \heatcellB{0.730} 
    & \heatcellB{0.700} \\
    
    OpenCLIP ConvNeXt-L (D, 320px)        
    & \heatcellB{0.130} 
    & \heatcellB{0.750} 
    & \heatcellB{0.780} 
    & \heatcellB{0.790} \\

        \midrule
    \textbf{Mean}
    & \heatcellB{0.156} 
    & \heatcellB{0.653} 
    & \heatcellB{0.692} 
    & \heatcellB{0.693} \\

    \bottomrule
  \end{tabular}
  \label{tab:ava_modelwise}
\end{table}

\textbf{KonIQ-10k}, introduced in \cite{hosu2020koniq10k}, is another aesthetics or image quality assessment (IQA) dataset that aims to model naturally occurring image distortions with mean opinion scores ranging roughly between 1 and 100. We use the official train-test splits. Results can be found in \cref{tab:koniq10k_modelwise}.

\begin{table}[H]
  \centering
  \small
  \setlength{\tabcolsep}{6pt}
  \renewcommand{\arraystretch}{1.2}
  \caption{
  \textbf{KonIQ-10k (Image Aesthetics)}. Spearman’s rank correlation $\rho$ across evaluation strategies.}
  \vspace{6pt}
  \begin{tabular}{lcccc}
    \toprule
    \textbf{Model} 
    & \makecell{\textbf{No-train}\\lower bound} 
    & \makecell{\textbf{Rankability}\\main} 
    & \makecell{\textbf{Nonlinear}\\upper bound} 
    & \makecell{\textbf{Finetuned}\\upper bound} \\
    \midrule
    ResNet-50                      
    & \heatcellB{0.563} 
    & \heatcellB{0.739} 
    & \heatcellB{0.739} 
    & \heatcellB{0.874} \\
    
    ViT-B/32                       
    & \heatcellB{0.488} 
    & \heatcellB{0.713} 
    & \heatcellB{0.753} 
    & \heatcellB{0.813} \\
    
    ConvNeXtV2-L                  
    & \heatcellB{0.487} 
    & \heatcellB{0.744} 
    & \heatcellB{0.765} 
    & \heatcellB{0.930} \\
    
    DINOv2 ViT-B/14                
    & \heatcellB{0.324} 
    & \heatcellB{0.681} 
    & \heatcellB{0.753} 
    & \heatcellB{0.948} \\
    
    OpenAI CLIP ResNet-50                 
    & \heatcellB{0.400} 
    & \heatcellB{0.800} 
    & \heatcellB{0.840} 
    & \heatcellB{0.900} \\
    
    OpenAI CLIP ViT-B/32                  
    & \heatcellB{0.460} 
    & \heatcellB{0.790} 
    & \heatcellB{0.830} 
    & \heatcellB{0.890} \\
    
    OpenCLIP ConvNeXt-L (D, 320px)        
    & \heatcellB{0.320} 
    & \heatcellB{0.860} 
    & \heatcellB{0.870} 
    & \heatcellB{0.950} \\

        \midrule
    \textbf{Mean}
    & \heatcellB{0.435} 
    & \heatcellB{0.761} 
    & \heatcellB{0.793} 
    & \heatcellB{0.901} \\
    
    \bottomrule
  \end{tabular}
  \label{tab:koniq10k_modelwise}
\end{table}

\subsection{Image recency}

\textbf{Historical Color Images (HCI)}, introduced in \cite{palermo2012dating}, was designed for the task of classifying an image by the decade during which it was taken. Therein emerges a natural ordering over the decades, defining the ordinal attribute of image \quotes{modernness} or \quotes{recency}. We use the split provided by \cite{yu2025rankingaware} in their repository (\href{https://github.com/uynaes/RankingAwareCLIP/tree/main/examples}{https://github.com/uynaes/RankingAwareCLIP/tree/main/examples}) and report the results in \cref{tab:hci_modelwise}.

\begin{table}[H]
  \centering
  \small
  \setlength{\tabcolsep}{6pt}
  \renewcommand{\arraystretch}{1.2}
  \caption{
  \textbf{HCI (Historical Color Images)}. Spearman’s rank correlation $\rho$ across evaluation strategies.}
  \vspace{6pt}
  \begin{tabular}{lcccc}
    \toprule
    \textbf{Model} 
    & \makecell{\textbf{No-train}\\lower bound} 
    & \makecell{\textbf{Rankability}\\main} 
    & \makecell{\textbf{Nonlinear}\\upper bound} 
    & \makecell{\textbf{Finetuned}\\upper bound} \\
    \midrule
    ResNet-50                      
    & \heatcellB{0.351} 
    & \heatcellB{0.600} 
    & \heatcellB{0.592} 
    & \heatcellB{0.614} \\
    
    ViT-B/32                       
    & \heatcellB{0.377} 
    & \heatcellB{0.592} 
    & \heatcellB{0.618} 
    & \heatcellB{0.529} \\
    
    ConvNeXtV2-L                  
    & \heatcellB{0.362} 
    & \heatcellB{0.631} 
    & \heatcellB{0.663} 
    & \heatcellB{0.771} \\
    
    DINOv2 ViT-B/14                
    & \heatcellB{0.131} 
    & \heatcellB{0.571} 
    & \heatcellB{0.601} 
    & \heatcellB{0.748} \\
    
    OpenAI CLIP ResNet-50                 
    & \heatcellB{0.320} 
    & \heatcellB{0.780} 
    & \heatcellB{0.770} 
    & \heatcellB{0.760} \\
    
    OpenAI CLIP ViT-B/32                  
    & \heatcellB{0.430} 
    & \heatcellB{0.770} 
    & \heatcellB{0.780} 
    & \heatcellB{0.760} \\
    
    OpenCLIP ConvNeXt-L (D, 320px)        
    & \heatcellB{0.300} 
    & \heatcellB{0.820} 
    & \heatcellB{0.790} 
    & \heatcellB{0.870} \\

        \midrule
    \textbf{Mean}
    & \heatcellB{0.324} 
    & \heatcellB{0.680} 
    & \heatcellB{0.688} 
    & \heatcellB{0.722} \\

    \bottomrule
  \end{tabular}
  \label{tab:hci_modelwise}
\end{table}

\section{Comparison with SOTA}
\label{supp:sota}

Prior research has presented results from dedicated or general efforts to solve the datasets considered in our study. Our main aim is to understand the rankability emerging out of the structure in visual embedding spaces, and we contextualize our numbers using reference metrics (lower bound provided by the no-encoder baseline and upper bound provided by nonlinear regression and finetuned encoders). However, in \cref{tab:rotated_rankability_table}, we also provide comparisons with state-of-the-art results to further contextualize our results. 
Some comparisons suggest that simple linear regression over pretrained embeddings often performs comparably with or even surpasses dedicated efforts. Although architectural and training dataset differences mean that this comparison is not always fair, we emphasize the contrast in implementational simplicty between dedicated efforts and simple linear regression over pretrained embeddings that are often readily available and easy to use. 

\begin{landscape}
{
\fontsize{8.5pt}{10pt}\selectfont
\renewcommand{\arraystretch}{1.1}
\setlength{\tabcolsep}{3pt}

\begin{longtable}{
  >{\centering\arraybackslash}m{1.8cm}  
  >{\centering\arraybackslash}m{1.8cm}  
  l
  c
  c
  p{4.5cm}
  p{4.5cm}
  p{4.5cm}
}

\caption{
\textbf{Comparing linear / nonlinear regression against recent state-of-the-art methods.}
“Linear” and “Nonlinear” use regression over CLIP-ConvNeXt-L embeddings. Dashes indicate metrics unreported in prior work. We take the numbers for age, aesthetics and recency from \cite{yu2025rankingaware}, and crowd count from \cite{ma2025clipebc}.
Under \quotes{Downstream model}, we report the components used on top of pretrained visual embeddings (CLIP or non-CLIP models); sometimes, we also report if the encoder itself was retrained.
Under \quotes{Downstream data}, we report the additional data used for training the method. Finally, under \quotes{Other ingredients}, we also report miscellaneous extra components used by the corresponding method. 
All method interpretations are to the best of our knowledge.
\label{tab:rotated_rankability_table}
} \\

\toprule
\textbf{Attribute} &
\textbf{Dataset} &
\textbf{Method} &
\textbf{Spearman $\rho$} &
\textbf{MAE} &
\textbf{Downstream model} &
\textbf{Downstream data} &
\textbf{Other ingredients} \\
\midrule
\endfirsthead

\multicolumn{8}{c}%
{\tablename\ \thetable\ -- \textit{Continued from previous page}} \\
\toprule
\textbf{Attribute} &
\textbf{Dataset} &
\textbf{Method} &
\textbf{Spearman $\rho$} &
\textbf{MAE} &
\textbf{Downstream model} &
\textbf{Downstream data} &
\textbf{Other ingredients} \\
\midrule
\endhead

\midrule
\multicolumn{8}{r}{\textit{Continued on next page}} \\
\endfoot

\bottomrule
\endlastfoot

\multirow[t]{4}{*}{Age} &
\multirow[t]{4}{*}{UTKFace} &
Yu et al.~\cite{yu2025rankingaware} &
-- &
3.83 &
Cross-attn encoder, two ranking heads, learnable text prompt tokens &
Images with age labels &
Text encoder \\
& &
MiVOLO \cite{kuprashevich2023mivolo} &
-- &
4.23 &
Regression heads &
Body images, face patches, age labels &
Feature enhancer module for fused joint representations \\
& &
Linear &
-- &
4.25 &
Linear regressor &
Images with age labels &
None \\
& &
Nonlinear &
-- &
4.10 &
2-layer MLP regressor &
Images with age labels &
None \\
\midrule

\multirow[t]{4}{*}{Age} &
\multirow[t]{4}{*}{Adience} &
Yu et al.~\cite{yu2025rankingaware} &
-- &
0.36 (0.03) &
Cross-attn encoder, two ranking heads, learnable text prompt tokens &
Images with age labels &
Text encoder \\
& &
OrdinalCLIP \cite{li2022ordinalclip} &
-- &
0.47 (0.06) &
(Retrain image encoder for the task) &
Images with age labels &
Text encoder; learn \quotes{continuous} rank prototype (text) embeddings for each rank \\
& &
Linear regressor &
-- &
0.48 (0.02) &
Linear &
Images with age labels &
None \\
& &
Nonlinear &
-- &
0.45 (0.02) &
2-layer MLP regressor &
Images with age labels &
None \\
\midrule

\multirow[t]{4}{*}{Crowd Count} &
\multirow[t]{4}{*}{UCF-QNRF} &
CLIP-EBC~\cite{ma2025clipebc} &
-- &
80.3 &
Blockwise classification module &
Images with count labels &
Text encoder \\
& &
CrowdCLIP~\cite{liang2023crowdclip} &
-- &
283.3 &
(Retrain image encoder)  &
Crowd images &
Text encoder, three-stage progressive filtering during inference \\
& &
Linear &
-- &
246.4 &
Linear &
Images with count labels &
None \\
& &
Nonlinear &
-- &
248.0 &
2-layer MLP &
Images with count labels &
None \\
\midrule

\multirow[t]{4}{*}{Crowd Count} &
\multirow[t]{4}{*}{ST-A} &
CLIP-EBC~\cite{ma2025clipebc} &
-- &
52.5 &
Blockwise classification module &
Images with count labels &
Text encoder \\
& &
CrowdCLIP~\cite{liang2023crowdclip} &
-- &
146.1 &
(Retrain image encoder) &
Crowd images &
Text encoder, three-stage progressive filtering during inference\\
& &
Linear &
-- &
167.1 &
Linear &
Images with count labels &
None \\
& &
Nonlinear &
-- &
151.7 &
2-layer MLP &
Images with count labels &
None \\
\midrule

\multirow[t]{4}{*}{Crowd Count} &
\multirow[t]{4}{*}{ST-B} &
CLIP-EBC~\cite{ma2025clipebc} &
-- &
6.6 &
Blockwise classification module &
Images with count labels &
Text encoder \\
& &
CrowdCLIP~\cite{liang2023crowdclip} &
-- &
69.3 &
(Retrain image encoder) &
Crowd images &
Text encoder, three-stage progressive filtering during inference \\
& &
Linear &
-- &
34.7 &
Linear &
Images with count labels &
None \\
& &
Nonlinear &
-- &
29.7 &
2-layer MLP &
Images with count labels &
None \\
\midrule


\multirow[t]{4}{*}{Aesthetics} &
\multirow[t]{4}{*}{AVA*} &
Yu et al.~\cite{yu2025rankingaware} &
0.747 &
-- &
Cross-attn encoder, two ranking heads, learnable text prompt tokens &
Images with MOS labels (1–10) &
Text encoder \\
& &
CLIP-IQA~\cite{wang2023exploring} &
0.415 &
-- &
Softmax over two similarity scores &
None &
Text encoder, prompt engineering, remove position embedding\\
& &
Linear &
0.749 &
-- &
Linear &
Images with MOS labels(1–10) &
None \\
& &
Nonlinear &
0.775 &
-- &
2-layer MLP &
Images with MOS labels (1–10) &
None \\
\midrule

\multirow[t]{4}{*}{Aesthetics} &
\multirow[t]{4}{*}{KonIQ-10k} &
Yu et al.~\cite{yu2025rankingaware} &
0.911 &
-- &
Cross-attn encoder, two ranking heads, learnable text prompt tokens &
Images with MOS labels (1–100) &
Text encoder \\
& &
CLIP-IQA~\cite{wang2023exploring} &
0.727 &
-- &
Softmax over two similarity scores &
Images with MOS labels (1–100) &
Text encoder, prompt engineering, remove position embedding \\
& &
Linear &
0.860 &
-- &
Linear &
Images with MOS labelss (1–100) &
None \\
& &
Nonlinear &
0.870 &
-- &
2-layer MLP &
Images with MOS labels (1–100) &
None \\
\midrule

\multirow[t]{4}{*}{Recency} &
\multirow[t]{4}{*}{HCI*} &
Yu et al.~\cite{yu2025rankingaware} &
-- &
0.32 (0.03) &
Cross-attn encoder, two ranking heads, learnable text prompt tokens &
Images with decade labels &
Text encoder \\
& &
OrdinalCLIP~\cite{li2022ordinalclip} &
-- &
0.67 (0.03) &
(Retrain image encoder for the task) &
Images with decade labels &
Text encoder; learn \quotes{continuous} rank prototype (text) embeddings for each rank \\
& &
Linear &
-- &
0.64 &
Linear &
Images with decade labels &
None \\
& &
Nonlinear &
-- &
0.60 &
2-layer MLP &
Images with decade labels &
None \\

\end{longtable}
}
\end{landscape}

\vspace{-1em}

\end{document}